\documentclass{article}

\usepackage{lmodern}
\usepackage{times}
\usepackage{soul}
\usepackage{url}
\usepackage[utf8]{inputenc}
\usepackage[small]{caption}
\usepackage{graphicx}
\usepackage{amsmath,amssymb,amsthm}
\usepackage{amsfonts}       
\usepackage{mathtools}
\usepackage{bm} 
\usepackage{booktabs}
\usepackage{algorithm}
\usepackage{algorithmic}
\usepackage[switch]{lineno}
\usepackage{multirow}
\usepackage{multicol}
\usepackage{subfigure}
\usepackage{booktabs}
\usepackage{makecell}
\usepackage{xcolor}

\usepackage[accepted]{icml2026}

\theoremstyle{plain}
\newtheorem{theorem}{Theorem}[section]
\newtheorem{proposition}[theorem]{Proposition}

\theoremstyle{definition}

\newtheorem{assumption}[theorem]{Assumption}
\theoremstyle{remark}

\usepackage[textsize=tiny]{todonotes}

\icmltitlerunning{DP-aware AdaLN-Zero: Taming Conditioning-Induced Heavy-Tailed Gradients in Differentially Private Diffusion}

\begin{document}

\twocolumn[
  \icmltitle{DP-aware AdaLN-Zero: Taming Conditioning-Induced Heavy-Tailed Gradients in Differentially Private Diffusion}




  \begin{icmlauthorlist}
    \icmlauthor{Tao Huang}{yyy}
    \icmlauthor{Jiayang Meng}{comp}
    \icmlauthor{Xu Yang}{yyy}
    \icmlauthor{Chen Hou}{yyy}
    \icmlauthor{Hong Chen}{comp}
  \end{icmlauthorlist}

  \icmlaffiliation{yyy}{School of Computer Science and Big Data, Minjiang University, Fuzhou, Fujian, China.}
  \icmlaffiliation{comp}{School of Information, Renmin University of China, Beijing, China}

  \icmlkeywords{Machine Learning, ICML}

  \vskip 0.3in
]




\begingroup
\setcounter{footnote}{0}
\renewcommand{\thefootnote}{}
\renewcommand{\theHfootnote}{notice.\arabic{footnote}}
\printAffiliationsAndNotice{}
\endgroup

\begin{abstract}
  Condition injection enables diffusion models to generate context-aware outputs, which is essential for many time-series tasks. However, heterogeneous conditional contexts (e.g., observed history, missingness patterns or outlier covariates) can induce heavy-tailed per-example gradients. Under \textbf{D}ifferentially \textbf{P}rivate \textbf{S}tochastic \textbf{G}radient \textbf{D}escent (DP-SGD), these rare conditioning-driven heavy-tailed gradients disproportionately trigger global clipping, resulting in outlier-dominated updates, larger clipping bias, and degraded utility under a fixed privacy budget. In this paper, we propose DP-aware AdaLN-Zero, a drop-in sensitivity-aware conditioning mechanism for conditional diffusion transformers that limits conditioning-induced gain without modifying the DP-SGD mechanism. DP-aware AdaLN-Zero jointly constrains conditioning representation magnitude and AdaLN modulation parameters via bounded re-parameterization, suppressing extreme gradient tail events before gradient clipping and noise injection. Empirically, DP-SGD equipped with \textbf{DP-aware AdaLN-Zero} improves interpolation/imputation and forecasting under matched privacy settings. We observe consistent gains on a real-world power dataset and two public ETT benchmarks over vanilla DP-SGD. Moreover, gradient diagnostics attribute these improvements to conditioning-specific tail reshaping and reduced clipping distortion, while preserving expressiveness in non-private training. Overall, these results show that sensitivity-aware conditioning can substantially improve private conditional diffusion training without sacrificing standard performance.

\end{abstract}

\section{Introduction}
\label{sec:intro}

Diffusion models learn expressive data distributions through iteratively denoising corrupted samples, and have become a leading paradigm for conditional generation \cite{ho2020ddpm,song2021score,fu2024unveil,zhan2024conditional}. In time-series domain, diffusion-based methods achieve strong performance in probabilistic forecasting and imputation by conditioning on historical context, covariates, or partially observed trajectories \cite{rasul2021timegrad,tashiro2021csdi}. However, such reliance on rich conditioning signals often involves sensitive attributes and individual-level histories, motivating mechanisms that enable model or synthetic data release while limiting inference about any single record.

Differential privacy (DP) provides a rigorous guarantee that the output of an algorithm changes only slightly across adjacent datasets differing in a single training example \cite{dwork2014foundations}. Differentially Private Stochastic Gradient Descent (DP-SGD) \cite{abadi2016dpsgd} is the predominant algorithm for differentially private training of deep models. It clips per-example gradients and adds calibrated Gaussian noise to ensure $(\varepsilon,\delta)$-DP. Yet, for diffusion models—particularly conditional diffusion—DP-SGD often yields a severe privacy-utility trade-off. Recent work mitigates this gap by tailoring parameterization, optimization, and sampling to diffusion \cite{dockhorn2023dpdm}, by leveraging public pretraining followed by differentially private fine-tuning \cite{ghalebikesabi2023useful}; or by reducing redundant privacy noise via reuse of forward-process noise \cite{wang2024dppromise}. However, a critical challenge remains unaddressed: conditioning amplifies gradient sensitivity under DP-SGD.

We study conditional diffusion for time-series data, where the conditioning variables include observed history, missingness patterns, or outlier covariates. The resulting conditioning distribution is often highly heterogeneous, driven by diverse missingness structures, rare events, and extreme covariate values. This heterogeneity can produce heavy-tailed per-example gradient norms, particularly along the conditioning pathway. Under DP-SGD, a small fraction of updates with unusually large conditioning-induced gradients can dominate the clipping criterion, forcing aggressive clipping. This increases the effective noise-to-signal ratio and introduces a systematic optimization bias. Namely, parameter updates become disproportionately influenced by rare conditioning outliers rather than representative examples, which degrades utility even when the model backbone would otherwise train stably. Importantly, this failure mode cannot be addressed by improving diffusion-specific DP mechanisms alone; it requires explicitly sensitivity-aware conditioning mechanisms.

This paper focuses specifically on conditional diffusions, which typically incorporate conditioning via widely used adaptive LayerNorm (AdaLN) modulation and its zero-initialized variant (AdaLN-Zero). While effective, such modulation can amplify sensitivity \cite{peebles2023dit}. To overcome this issue, we propose \textbf{DP-aware AdaLN-Zero}, which preserves the DP mechanism while reshaping per-example gradient norms to suppress conditioning-induced spikes, reduce outlier-dominated clipping, and stabilize training under a fixed privacy budget. Specifically, our contributions are summarized as follows:

\begin{itemize}
    \item We identify a conditioning-driven sensitivity imbalance in differentially private conditional diffusion models: rare conditioning events can induce heavy-tailed per-example gradients that disproportionately trigger gradient clipping;
    \item We propose DP-aware AdaLN-Zero, a sensitivity-aware conditioning mechanism that bounds AdaLN modulation by constraining conditioning representation magnitude and AdaLN modulation parameters;
    \item Empirically, our approach stabilizes private training dynamics and yields higher downstream utility at the same noise scale than vanilla DP-SGD.
\end{itemize}

\section{Related Work}
\label{sec:related-work}

\paragraph{Diffusion models and conditional generation.}
Diffusion models learn to reverse a corruption process, enabling high-quality generation \cite{sohl2015deep,ho2020ddpm,song2021score}. In conditional diffusion, guidance techniques offer additional control via contextual information (e.g., labels, or observed history) at inference time. Classifier guidance improves conditional alignment by exploiting gradients from an auxiliary classifier \cite{dhariwal2021diffusion,ho2022cfg}. More recently, diffusion transformers (DiT) popularized conditioning via normalization-based modulation mechanisms such as AdaLN and AdaLN-Zero \cite{peebles2023dit}. While effective, modulation-based conditioning can amplify the gain of the conditioning pathway. Amplified responses can result in disproportionately large per-example gradients, making clipping more aggressive under DP-SGD.

\paragraph{Diffusion models for time series.}
Diffusion models have been adapted for time-series forecasting and imputation by conditioning denoisers on temporal context and partially observations. TimeGrad introduces an autoregressive diffusion framework for multivariate probabilistic forecasting \cite{rasul2021timegrad}, and CSDI trains conditional score-based diffusion for probabilistic time-series imputation \cite{tashiro2021csdi}. However, they do not consider differentially private training, and their conditioning signals (e.g., observed masks/values) can be highly heterogeneous—exactly the regime in which DP-SGD is vulnerable to heavy-tailed per-example gradients.

\paragraph{Differential privacy for deep learning.}
DP formalizes privacy as the stability of an algorithm’s output to changes in any single training record \cite{dwork2014foundations}. DP-SGD enforces this guarantee by clipping per-example gradients and adding calibrated Gaussian noise \cite{abadi2016dpsgd}. However, frequent clipping introduces bias and reduces the effective optimization signal \cite{11321288}. This effect is amplified when gradient norms are heavy-tailed or when particular submodules occasionally produce extreme gradients.

\paragraph{Differentially private diffusion models.}
DPDM \cite{dockhorn2023dpdm} demonstrates that diffusion models can generate high-quality samples under DP by leveraging diffusion-specific design choices, including tailored sampling procedures and adaptations of DP-SGD. DP-promise \cite{wang2024dppromise} further argues that naively applying DP-SGD to diffusion can inject redundant gradient noise on top of the stochasticity already present in the forward process, and proposes reusing forward-process noise to obtain approximate DP with improved utility. However, these approaches do not readily extend to larger-scale diffusion models with explicit conditional pathways, where conditioning-induced gradient amplification can dominate the privacy–utility trade-off.

\paragraph{Limitations in differentially private diffusion work.}
Despite recent progress, existing differentially private diffusion methods largely optimize global training components, such as privacy accounting, sampler design, and pretraining/fine-tuning. They are evaluated mainly on unconditional or weakly conditional image generation. Moreover, they do not explicitly address a failure mode in conditional diffusion models: conditioning-induced sensitivity imbalance. In conditional time-series diffusion, conditioning can produce rare but extreme per-example gradients. Under DP-SGD, these extremes disproportionately trigger clipping, causing outlier-dominated updates, systematic bias toward small-norm directions, and reduced effective signal-to-noise after noise injection. Crucially, better samplers or global noise reallocation do not directly prevent conditioning from acting as an amplification channel that creates heavy-tailed gradient norms.

\section{Method}\label{section_three}

\subsection{Preliminaries}

\paragraph{Conditional Diffusion.}
We denote the conditioning vector by $\mathbf{c}\in\mathbb{R}^k$. Let $g$ be gradients: $g_i$ is the per-example gradient, with $\|g_i\|_2$ its $\ell_2$ norm. We consider a conditional time-series diffusion model parameterized by $\theta$, whose denoiser predicts noise (or velocity) from an input $x$ conditioned on $\mathbf{c}$ (e.g., covariates, mask-aware statistics, or global context):
\begin{equation}
    y = f_\theta(x, \mathbf{c}),
\end{equation}
where $f_\theta$ is a stack of AdaLN-Zero blocks. For a hidden state $h \in \mathbb{R}^d$ and a condition $\mathbf{c} \in \mathbb{R}^k$, a typical AdaLN-Zero block computes
\begin{equation}
    u = \mathrm{LN}(h),\,v = \gamma \odot u + \beta,\,h = F(v;\theta_F),\,y = x + \alpha \odot h,
\end{equation}
where $\mathrm{LN}$ is LayerNorm, $F$ is a self-attention/MLP subnetwork, and $(\gamma,\beta,\alpha)$ are per-block modulation parameters obtained from $\mathbf{c}$. This conditioning pathway acts as a gain control: large modulation values amplify activations and local Jacobians, yielding rare but extreme per-example gradients.

\paragraph{Vanilla DP-SGD.}
Consider training with vanilla DP-SGD, for a mini-batch $D = \{z_i\}_{i=1}^B$ with examples $z_i=(x_i,\mathbf{c}_i)$, let
\begin{equation}
    g_i(\theta) \;=\; \nabla_\theta \,\ell\!\big(f_\theta(z_i)\big)
\end{equation}
denote the per-example gradient of the training loss $\ell$. Vanilla DP-SGD clips each per-example gradient to a threshold $C>0$:
\begin{equation}
    \tilde{g}_i(\theta)=
    g_i(\theta)\cdot
    \min\!\left(1,\frac{C}{\|g_i(\theta)\|_2}\right),
\end{equation}
and releases a noisy average of clipped gradients. The standard analysis upper-bounds the $\ell_2$-sensitivity of this update by $\frac{C}{B}$, independent of model structure and conditioning.

\begin{figure*}[t]
    \centering
    \subfigure[\textbf{ECDF} of gradient norms in normal training.]{
        \includegraphics[width=0.88\textwidth]{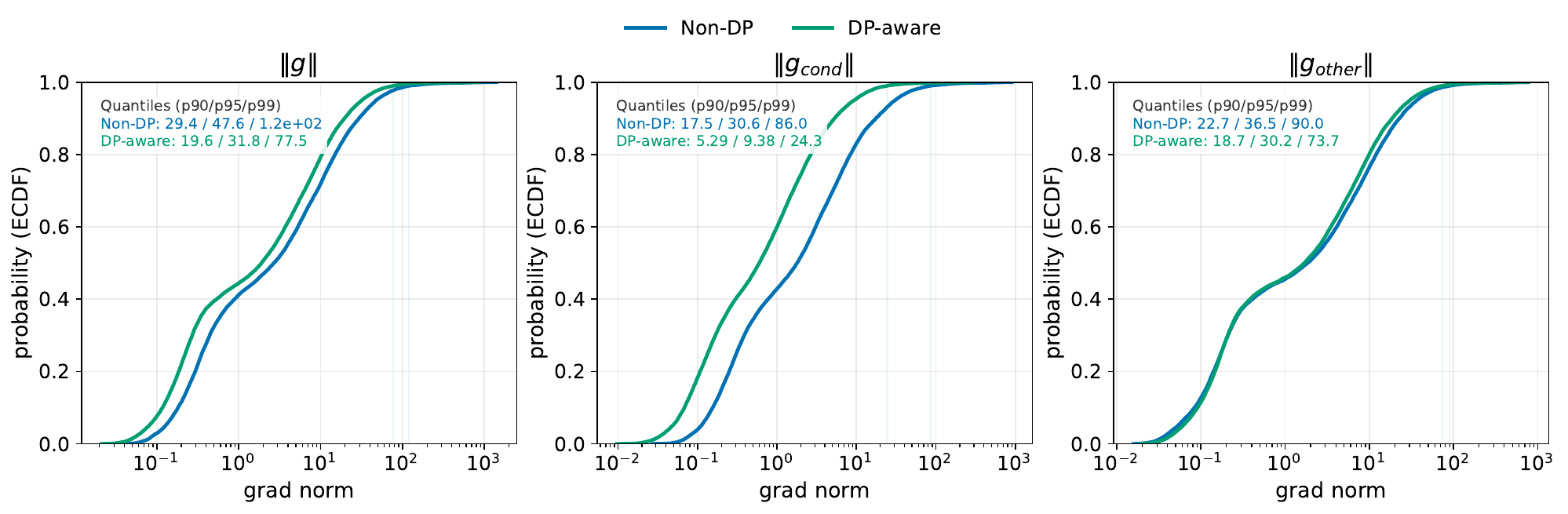}
        \label{fig:gradnorm:ecdf:no-dp}
    }
    \subfigure[\textbf{CCDF (tail)} of gradient norms in normal training.]{
        \includegraphics[width=0.88\textwidth]{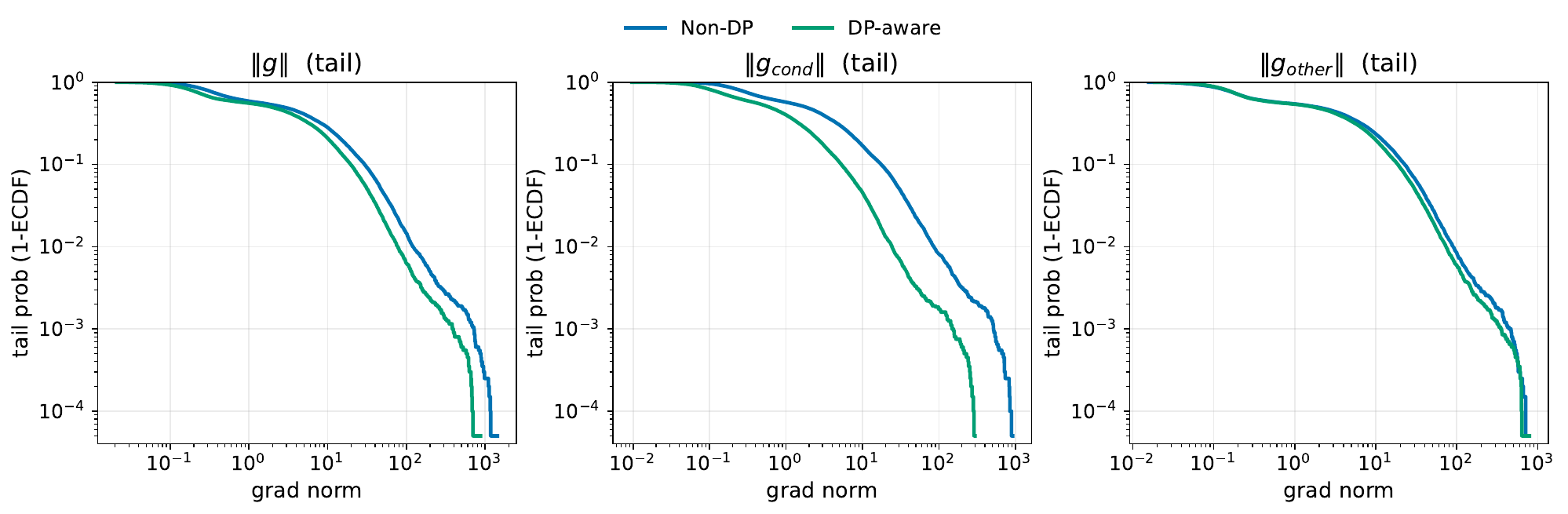}
        \label{fig:gradnorm:tail:no-dp}
    }
    \caption{\textbf{Condition-amplified extremes exist even without DP.}
We compare normal training with training under DP-aware constraints \emph{without} DP.
Under normal training, $\|g_{\mathrm{cond}}\|_2$ is comparable to $\|g_{\mathrm{other}}\|_2$ at typical quantiles
(see the blue curve in Figure~\ref{fig:gradnorm:ecdf:no-dp}),
but $\|g_{\mathrm{cond}}\|_2$ exhibits rarer and heavier high-end tail events
(see the blue curve in Figure~\ref{fig:gradnorm:tail:no-dp}).
In contrast, DP-aware constraints \emph{selectively} suppress the high-end tail of $\|g_{\mathrm{cond}}\|_2$
(and consequently that of $\|g\|_2$) far more than $\|g_{\mathrm{other}}\|_2$:
$p99$ drops by $\sim 3.5\times$ for $\|g_{\mathrm{cond}}\|_2$ vs.\ $\sim 1.2\times$ for $\|g_{\mathrm{other}}\|_2$.
This indicates targeted suppression of conditioning-induced amplification rather than uniform shrinkage,
and suggests that (in DP-SGD) clipping events would be disproportionately governed by rare conditioning-path extremes.}
    \label{fig:grad_norm_distribution_no-dp}
\end{figure*}
\vspace{-4mm}

\subsection{Structural Limits of Global Clipping}

In AdaLN-Zero conditional diffusion, employing a single global clipping threshold $C$ fails to account for a crucial architectural asymmetry. To address this, we partition the parameters into two distinct groups: (i) conditioning-path parameters $\theta_{\mathrm{cond}}$ (the projections that map $\mathbf{c}$ to $(\gamma,\beta,\alpha)$ and parameters directly on the conditioning injection pathway); and (ii) other parameters $\theta_{\mathrm{other}}$ (all remaining parameters). The corresponding per-example gradient decomposes as:
\[
\begin{aligned}
    g_i=\big(g^{\mathrm{cond}}_i, g^{\mathrm{other}}_i\big).
\end{aligned}
\]

This decomposition highlights two coupled issues:

\emph{(i) Clipping is often governed by rare conditioning-path tail events.}
Due to multiplicative modulation in AdaLN-Zero, the conditioning map $\mathbf{c}\mapsto(\gamma,\beta,\alpha)$ can induce heavy-tailed gradients on $\theta_{\mathrm{cond}}$. Equivalently, in the high-threshold regime relevant to clipping, conditioning-driven extremes are more likely to exceed large thresholds for sufficiently large $t$:
\[
\Pr\!\big(\|g^{\mathrm{cond}}_i\|_2 > t\big) \;\gg\; \Pr\!\big(\|g^{\mathrm{other}}_i\|_2 > t\big).
\]
Thus, rare condition-amplified gradient spikes can disproportionately elevate $\|g_i\|_2$ beyond $C$, thereby increasing the activation probability of gradient clipping.

\emph{(ii) Global clipping uses a single scalar shrinkage, attenuating conditional learning.}
DP-SGD clips by rescaling each per-example gradient:
$\tilde g_i = \eta_i g_i$, where $\eta_i=\min\!\left(1,\frac{C}{\|g_i\|_2}\right)$.
When $\|g_i\|_2 > C$ is triggered primarily by a spike in $g^{\mathrm{cond}}_i$,
the scalar $\eta_i$ uniformly shrinks all coordinates,
including non-extreme updates in $\theta_{\mathrm{other}}$ and the non-spiking part of $\theta_{\mathrm{cond}}$.
This creates a privacy-utility dilemma:
decreasing $C$ reduces the DP noise scale (proportional to $C$) but increases the clipping-induced distortion, while increasing $C$ reduces clipping distortion but amplifies DP noise.

Figure~\ref{fig:grad_norm_distribution_no-dp} shows evidence for this issue in non-private training. The parameter set $\theta_{\mathrm{cond}}$ is often much smaller than $\theta_{\mathrm{other}}$. However, the per-example gradient norm $\|g_i^{\mathrm{cond}}\|_2$ has a much heavier high-end tail than $\|g_i^{\mathrm{other}}\|_2$. This result means that conditioning can create unusually large gradients even in standard training.

Selective bounds on conditioning gain reduce these extreme values. In particular, we attempt to shrink the high-end tail of $\|g_i^{\mathrm{cond}}\|_2$. As a result, the high-end tail of the total norm $\|g_i\|_2$ can be reduced. In DP-SGD, clipping depends on $\|g_i\|_2$ through $\eta_i$. A smaller tail in $\|g_i\|_2$ reduces clipping distortion and helps the model learn from the conditioning signal.

\subsection{DP-aware AdaLN-Zero Design}
\label{sec:dp_aware_design}

The discussion above identifies the conditioning pathway as the key target. The global condition $\mathbf{c}$ and modulation parameters $(\gamma,\beta,\alpha)$ control the magnitude of hidden representations and their Jacobians. We aim to bound this forward-pass gain, while leaving DP-SGD unchanged.

DP-aware AdaLN-Zero applies deterministic per-block constraints to bound the effect of $\mathbf{c}$ and the AdaLN modulation parameters on the conditioning pathway. For each block, we first $\ell_2$-bound the global condition:
\begin{equation}
    \hat{\mathbf{c}} \;=\; \mathrm{Proj}_{\|\mathbf{c}\|_2 \le c_{\max}}(\mathbf{c}),
    \label{eq:c_l2_clip}
\end{equation}
where $c_{\max}>0$ is a fixed constant. The AdaLN parameters are then obtained from $\hat{\mathbf{c}}$ through linear projections,
\begin{equation}(\gamma_{\mathrm{raw}},\beta_{\mathrm{raw}},\alpha_{\mathrm{raw}})=W\hat{\mathbf{c}} + b,
    \label{eq_6}
\end{equation}
followed by coordinatewise clipping:
\begin{equation}
(\gamma,\beta,\alpha)=\mathcal{B}_M\big((\gamma_{\mathrm{raw}},\beta_{\mathrm{raw}},\alpha_{\mathrm{raw}}),(\gamma_{\max},\beta_{\max},\alpha_{\max})\big),
    \label{eq:adaln_bounds}
\end{equation}
where $(c_{\max}, \gamma_{\max}, \beta_{\max}, \alpha_{\max})$ control per-block modulation strength by enforcing $|\gamma|\le \gamma_{\max}$, $|\beta|\le \beta_{\max}$, and $|\alpha|\le \alpha_{\max}$. By default, we implement the coordinate-wise bounding operator $\mathcal{B}_M(\cdot)$ as $\mathcal{B}_M^{\tanh}(x)=M\tanh(\frac{x}{M})$. 

The bound in Eq.(\ref{eq:adaln_bounds}) introduces negligible overhead and is applied before gradient computation. By limiting the magnitude of conditioning signal and the induced modulation parameters, it constrains intermediate activations and their Jacobians, suppressing rare condition-amplified spikes that would otherwise produce extreme per-example gradients.

DP-aware AdaLN-Zero improves DP-SGD without modifying DP mechanism. In vanilla DP-SGD (without DP-aware deterministic constraints), a conditioning-induced spike can push the global $\ell_2$-norm $\|g_i\|_2$ above the clipping threshold $C$, causing the entire gradient to be uniformly rescaled and thus attenuating conditional updates, even for parameters that do not contribute to the spike. By limiting conditioning-path amplification in every block via Eqs.(\ref{eq:c_l2_clip})-(\ref{eq:adaln_bounds}), our approach mitigates such spike-induced global shrinkage and the resulting clipping distortion.

\subsection{Sensitivity Analysis of DP-Aware AdaLN-Zero}
\label{sec:theory_sensitivity}

This section provides a structural worst-case analysis to justify why bounding the AdaLN-Zero conditioning pathway can control the magnitude of per-example gradients. Although stated in terms of $\ell_2$-sensitivity, the analysis should be interpreted as a mechanistic guarantee arising from architectural constraints.

We first derive a per-example gradient bound under DP-aware constraints. Assuming standard regularity conditions (bounded inputs, Lipschitz blocks on bounded domains, bounded loss gradient, bounded spectral norms), intermediate Jacobians remain controlled whenever activations are bounded. DP-aware constraints guarantee boundedness along the conditioning pathway. Specifically, for each block,
\begin{equation}
\begin{aligned}
\|\mathbf{c}\|_2 \le c_{\max},\,&
    \|\gamma\|_\infty \le \gamma_{\max},\,\\
    \|\beta\|_\infty \le \beta_{\max},\,&
    \|\alpha\|_\infty \le \alpha_{\max}.
    \label{eq:dp_aware_bounds_main}
\end{aligned}
\end{equation}
These bounds limit the modulation strength, thereby controlling hidden states and their Jacobians.

Let $\theta$ be all model parameters and $z=(x,\mathbf{c})$ a training example. We write $\nabla_\theta \ell\big(f_\theta(x,\mathbf{c})\big)$ for the corresponding per-example gradient. For analysis, we partition $\theta$ into parameters in $F$ (attention/MLP), projection layers generating $(\gamma,\beta,\alpha)$ from $\mathbf{c}$, and remaining parameters (e.g., LayerNorm, skip connections). Under Eq.(\ref{eq:dp_aware_bounds_main}), each component’s contribution to gradient norm can be bounded, yielding Proposition \ref{prop:gradient_bound_main}.

\begin{proposition}[Per-Example Gradient Bound with DP-aware Constraints]
\label{prop:gradient_bound_main}
Under the DP-aware constraints (Eq.(\ref{eq:dp_aware_bounds_main})) and the standard regularity assumptions, there exist non-negative, architecture-dependent constants $A_0,a_c,a_\gamma,a_\beta,a_\alpha \ge 0$, such that for every training example $z=(x,\mathbf{c})$,
\begin{equation}
\begin{aligned}
    \bigl\| \nabla_\theta \ell\big(f_\theta(x,\mathbf{c}) \big) \bigr\|_2 \le
    S_{\mathrm{aware}}, \quad\quad\quad\quad\quad\\
    S_{\mathrm{aware}}\coloneqq
    A_0
    + a_c c_{\max}
    + a_\gamma \gamma_{\max}
    + a_\beta \beta_{\max}
    + a_\alpha \alpha_{\max}.
\end{aligned}
    \label{eq:Saware_def_main}
\end{equation}
\end{proposition}
The proof of Proposition \ref{prop:gradient_bound_main} is given in Appendix ~\ref{appendix:proof_dp_aware}.

Next, we derive the sensitivity of the DP-SGD update and examine its implications for clipping. Consider one DP-SGD step with batch size $B$ and global clipping threshold $C$. For each example $z_i$ in a batch $D = \{z_i\}_{i=1}^B$, DP-SGD computes $\tilde g_i=g_i \cdot \min\!\left(1, \frac{C}{\|g_i\|_2}\right)$ and $q(D) = \frac{1}{B} \sum_{i=1}^B \tilde g_i$.

For vanilla DP-SGD, clipping ensures $\|\tilde g_i\|_2 \le C$, yielding the global $\ell_2$-sensitivity bound: $\Delta_2(q_{\mathrm{vanilla}}) \le \frac{C}{B}$.  For DP-SGD with DP-aware AdaLN-Zero, Proposition~\ref{prop:gradient_bound_main} implies $\|g_i\|_2 \le S_{\mathrm{aware}}$ for all $i$. When $S_{\mathrm{aware}} \le C$, clipping is never triggered and $\tilde g_i = g_i$ for all examples. Consequently, the sensitivity satisfies $\Delta_2(q_{\mathrm{aware}})\le \frac{S_{\mathrm{aware}}}{B}$.

To facilitate a direct comparison between the sensitivity of our method and that of vanilla DP-SGD, we define
\begin{equation}
    \rho
    \coloneqq
    \frac{\Delta_2(q_{\mathrm{aware}})}{\Delta_2(q_{\mathrm{vanilla}})}
    \le
    \frac{S_{\mathrm{aware}}}{C}.
    \label{eq:rho_basic_main}
\end{equation}
In our experiments, we match $(C,\sigma)$ between vanilla DP-SGD (DP-vanilla) and DP-SGD with DP-aware AdaLN-Zero (DP-aware) and focus on utility. Eq.(\ref{eq:rho_basic_main}) is adopted as a structural statement: by limiting conditioning-induced gain, DP-aware suppresses outliers and mitigates the distortion of conditional updates caused by clipping. In Appendix ~\ref{appendix:sensitivity_details}, we further show that under certain conditions, $\rho \leq 1$ always holds.

\subsection{Diagnostic of Gradient Dynamics}\label{sec:3.5}

\begin{table*}[t]
    \centering
    \resizebox{0.93\textwidth}{!}{
    \begin{tabular}{lc|ccc|ccc}
        \toprule
        & & \multicolumn{3}{c}{Interpolation/Imputation} & \multicolumn{3}{c}{ Forecasting} \\
        \cmidrule(lr){3-5} \cmidrule(lr){6-8}
        Model & $\sigma$ &
        \makecell{point\_RMSE $\downarrow$} &
        \makecell{point\_MAPE $\downarrow$} &
        \makecell{point\_MAE $\downarrow$} &
        \makecell{point\_RMSE $\downarrow$} &
        \makecell{dist\_JS $\downarrow$} &
        \makecell{temp\_spec\_dist $\downarrow$} \\
        \midrule

        Non-DP & - & 0.584 & 1.174\% & 0.476 & 0.208 & 0.767 & 1.484e-4 \\
        \midrule

        DP-vanilla & \multirow{2}{*}{0.03} & 3.034 & 5.673\% & 2.391 & 0.556 & 0.736 & 7.848e-4 \\
        DP-aware   & & \textbf{1.804} & \textbf{3.541\%} & \textbf{1.411} & \textbf{0.296} & \textbf{0.684} & \textbf{2.900e-4} \\
        \addlinespace

        DP-vanilla & \multirow{2}{*}{0.05} & 3.498 & 9.339\% & 3.110 & 0.567 & 0.643 & 1.650e-3 \\
        DP-aware   & & \textbf{2.019} & \textbf{4.970\%} & \textbf{1.987} & \textbf{0.423} & \textbf{0.636} & \textbf{8.558e-4} \\
        \addlinespace

        DP-vanilla & \multirow{2}{*}{0.1} & 5.787 & 11.674\% & 4.702 & 1.637 & 0.833 & 1.902e-2 \\
        DP-aware   & & \textbf{2.718} & \textbf{5.267\%} & \textbf{2.122} & \textbf{0.671} & \textbf{0.757} & \textbf{3.240e-3} \\
        \addlinespace

        DP-vanilla & \multirow{2}{*}{0.2} & 6.812 & 12.645\% & 5.148 & 1.646 & 0.794 & 1.592e-2 \\
        DP-aware   & & \textbf{4.689} & \textbf{8.406\%} & \textbf{3.442} & \textbf{1.262} & \textbf{0.732} & \textbf{1.052e-2} \\
        \bottomrule
    \end{tabular}}
    \caption{\textbf{Results on PrivatePower under matched DP training.} We evaluate mask-conditioned interpolation/imputation and forecasting. All metrics are lower-is-better. Bold marks the better of DP-vanilla and DP-aware at each $\sigma$.}
    \label{tab:main-results-extended}
\end{table*}
\vspace{-2mm}

Section \ref{sec:theory_sensitivity} derives a worst-case theoretical bound on per-example gradients for DP-aware AdaLN-Zero. However, such guarantees often fail to capture typical training behavior. We therefore complement the theory with empirical diagnostics of gradient distributions and clipping-induced distortion, offering a more faithful view of DP-SGD optimization dynamics.

Concretely, we quantify how DP-aware constraints affect typical per-example gradient magnitudes through an empirical factor $\rho_{\mathrm{emp}}$. We run two diagnostic trainings, DP-vanilla and DP-aware, under identical architecture, optimizer, and DP settings (global clipping threshold $C$ and noise multiplier $\sigma$). During training, for each sample $z_i=(x_i,\mathbf{c}_i)$, we log the full per-example gradient norm $\|g_i\|_2$, as well as the norms of its two parameter partitions: $\|g_i^{\mathrm{cond}}\|_2 =\|\nabla_{\theta_{\mathrm{cond}}}\ell(f_\theta(z_i))\|_2$ and $\|g_i^{\mathrm{other}}\|_2 = \|\nabla_{\theta_{\mathrm{other}}}\ell(f_\theta(z_i))\|_2$.

Since these norms fluctuate across training steps, we summarize them with robust statistics. Let $\mathrm{Stat}(\cdot)$ denote a robust summary operator—such as $p95/p99$—applied to all logged per-example gradients. We define $S_{\mathrm{total}}^{\text{(vanilla)}} \coloneqq \mathrm{Stat}\big(\|g_i\|_2\big)$ and $S_{\mathrm{total}}^{\text{(aware)}} \coloneqq \mathrm{Stat}\big(\|g_i\|_2\big)$, where the statistic is evaluated over the DP-vanilla and DP-aware diagnostic trajectories, respectively. The reduction factors are then
\begin{equation}
    \rho_{\mathrm{emp}}
    \coloneqq
    \frac{
        S_{\mathrm{total}}^{\text{(aware)}}
    }{
        S_{\mathrm{total}}^{\text{(vanilla)}}
    },\,
    \rho_{\mathrm{cond}}
    \coloneqq
    \frac{
        S_{\mathrm{cond}}^{\text{(aware)}}
    }{
        S_{\mathrm{cond}}^{\text{(vanilla)}}
    }.
    \label{eq:rho_emp_def}
\end{equation}
Here, $\rho_{\mathrm{emp}, \mathrm{cond}}$ captures the average reduction in per-example gradient-norm scale (or tail quantile) along training trajectories. It provides no worst-case sensitivity guarantee and is not assumed to upper-bound the theoretical $\rho$ in Eq.(\ref{eq:rho_basic_main}). We thus use $\rho_{\mathrm{emp}}$ only as a diagnostic metric for gradient dynamics.

\section{Experiments}\label{experiments}
\subsection{Experimental Setup}
\label{sec:experimental-setup}

\paragraph{Datasets.}
We evaluate our method on one real-world electricity dataset (PrivatePower) and two public benchmarks (ETTh1 and ETTm1). PrivatePower is an hourly single-meter electricity-usage time series spanning 2023-01-01 to 2025-08-22, with one record per hour. Each preprocessed window contains the standardized target \texttt{power\_usage} along with time-derived and user-derived conditioning channels. We set the maximum channel number to $K_{\max}=7$ and the sequence length to $L=168$ (one week). ETTh1 (hourly) and ETTm1 (15-min) are widely used multivariate benchmarks from the Electricity Transformer Temperature (ETT) suite~\cite{zhou2021informer}, comprising seven variables (one target and six covariates) and following standard chronological train/validation/test splits. See Appendix~\ref{app:exp-details} for additional dataset details.

\paragraph{Model.}
We implement the conditional diffusion model of \textsc{TimeDiT} and use its masking-based unified training scheme (random/stride/block masks) to support multiple conditional tasks (forecasting, interpolation, and imputation~\cite{cao2024timedit}). For PrivatePower, we use a Transformer backbone with 8 layers, hidden size 256, and 8 attention heads. For ETTh1/ETTm1, we keep the same architecture and adjust only $L_{\max}$ to match the dataset resolution and ETT evaluation window.

\begin{figure*}[t]
    \centering
    \subfigure[\textbf{ECDF} of per-example gradient norms in training at noise multiplier $\sigma=0.20$.]{
        \includegraphics[width=0.89\textwidth]{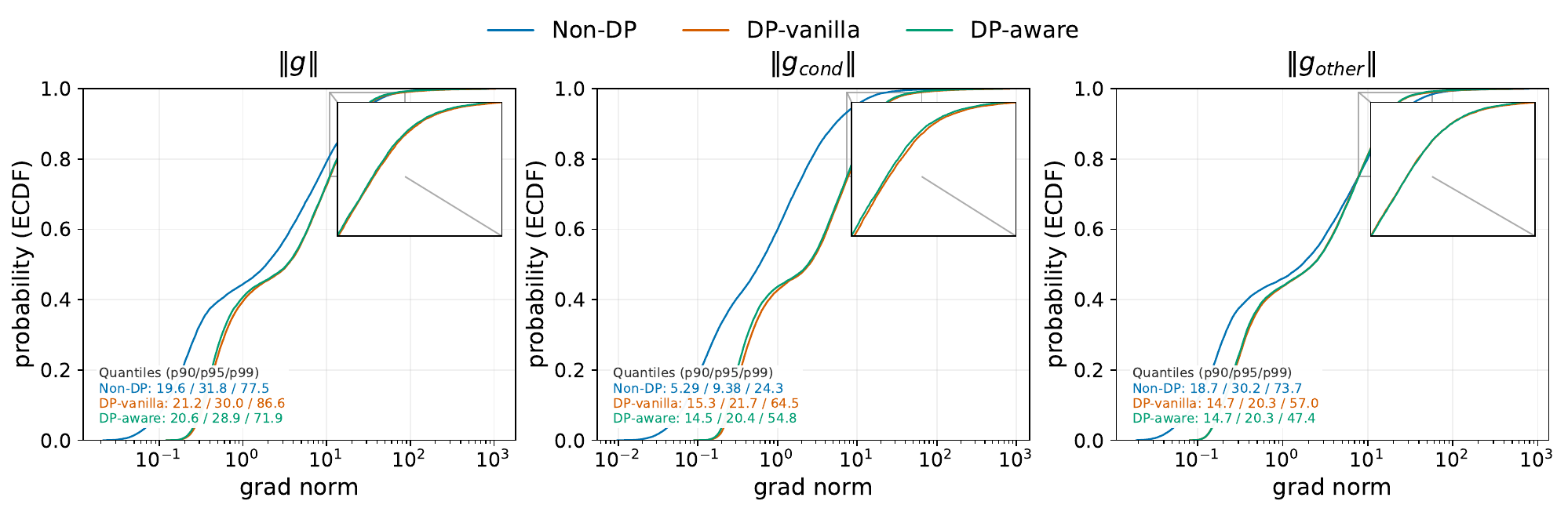}
        \label{fig:gradnorm:ecdf:0.20}
        }
    \subfigure[\textbf{CCDF (tail)} of per-example gradient norms at noise multiplier $\sigma=0.20$.]{
        \includegraphics[width=0.89\textwidth]{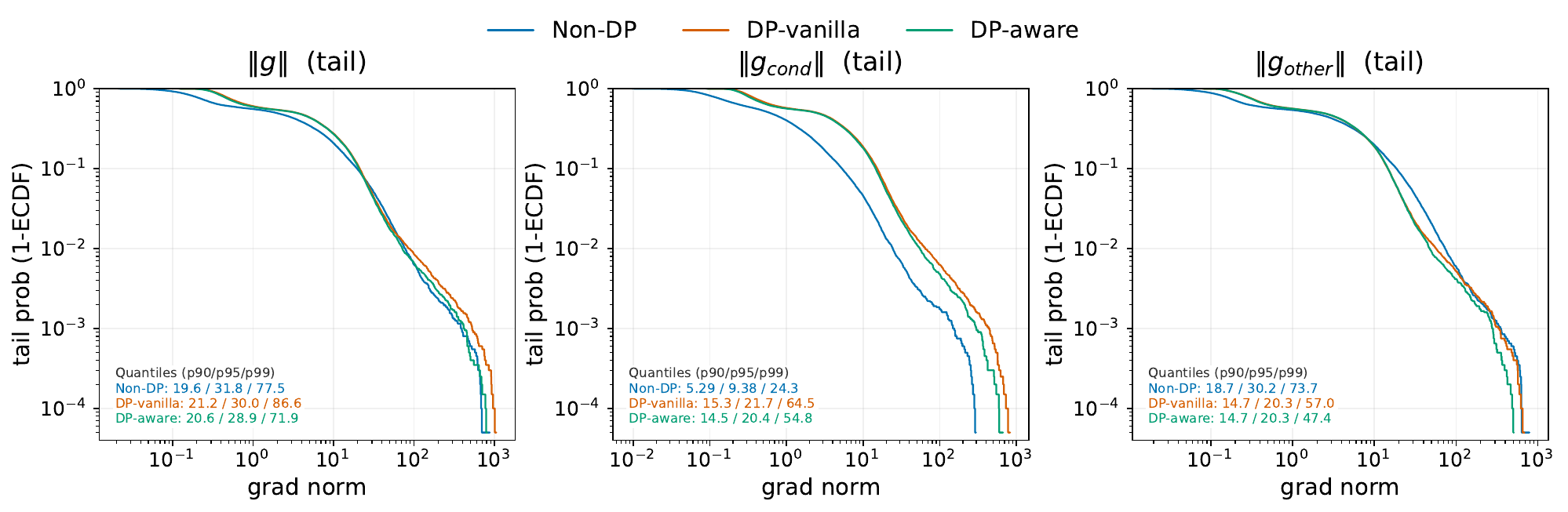}
        \label{fig:gradnorm:tail:0.20}
    }
    \caption{\textbf{Gradient-norm distributions under DP training}. Compared to DP-vanilla, DP-aware primarily suppresses the extreme tail of the conditioning pathway with minimal impact on the bulk of the distribution, indicating fewer condition-amplified outliers rather than uniform gradient shrinkage.}
    \label{fig:grad_norm_distribution}
\end{figure*}

\begin{table*}[t]
    \centering
    \resizebox{0.78\textwidth}{!}{
        \begin{tabular}{llcccccccccc}
            \toprule
            \multirow{2}{*}{\textbf{$\sigma$}} & \multirow{2}{*}{Model} &
            \multicolumn{5}{c}{p95 Statistics} & \multicolumn{5}{c}{p99 Statistics} \\
            \cmidrule(lr){3-7} \cmidrule(lr){8-12}
            & & $S_{\text{other}}$ & $S_{\text{cond}}$ & $S_{\text{total}}$ & $\rho_{\text{emp}}$ & $\rho_{\text{cond}}$
              & $S_{\text{other}}$ & $S_{\text{cond}}$ & $S_{\text{total}}$ & $\rho_{\text{emp}}$ & $\rho_{\text{cond}}$ \\
            \midrule
            \multirow{2}{*}{0.03} & DP-vanilla & 25.0 & 17.8 & 31.1 & -- & -- & 58.4 & 49.5 & 76.5 & -- & -- \\
                                 & DP-aware   & 24.6 & 16.4 & 29.7 & \textbf{0.955} & \textbf{0.921} & 57.5 & 43.7 & 72.0 & \textbf{0.949} & \textbf{0.883} \\
            \addlinespace[0.5em]
            \multirow{2}{*}{0.05} & DP-vanilla & 24.1 & 19.6 & 31.4 & -- & -- & 62.4 & 62.4 & 89.2 & -- & -- \\
                                 & DP-aware   & 24.2 & 18.9 & 30.9 & \textbf{0.984} & \textbf{0.964} & 55.6 & 55.5 & 77.3 & \textbf{0.867} & \textbf{0.889} \\
            \addlinespace[0.5em]
            \multirow{2}{*}{0.10} & DP-vanilla & 23.8 & 21.4 & 32.1 & -- & -- & 62.8 & 68.3 & 92.7 & -- & -- \\
                                 & DP-aware   & 24.4 & 20.3 & 31.2 & \textbf{0.972} & \textbf{0.949} & 54.9 & 57.4 & 83.1 & \textbf{0.896} & \textbf{0.840} \\
            \addlinespace[0.5em]
            \multirow{2}{*}{0.20} & DP-vanilla & 20.3 & 21.7 & 30.0 & -- & -- & 57.0 & 64.5 & 86.6 & -- & -- \\
                                 & DP-aware   & 20.3 & 20.4 & 28.9 & \textbf{0.963} & \textbf{0.940} & 47.4 & 54.8 & 71.9 & \textbf{0.830} & \textbf{0.850} \\
            \bottomrule
        \end{tabular}}
    \caption{\textbf{Gradient-norm diagnostics on PrivatePower across noise multipliers $\sigma$.} We report robust high-percentile summaries ($p95$/$p99$) of per-example gradient norms for conditioning-path parameters $\|g_{\mathrm{cond}}\|_2$ and remaining parameters $\|g_{\mathrm{other}}\|_2$. We also report a compact proxy $S_{\text{total}}$, empirical diagnostics $\rho_{\mathrm{emp}}$ and $\rho_{\mathrm{cond}}$; values below $1$ indicate reductions relative to DP-vanilla under matched $(C,\sigma)$.}
    \label{tab:grad-norms-multi-m}
\end{table*}
\vspace{-2mm}

\paragraph{Conditional tasks based masks.}
Let $x\in \mathbb{R}^{L\times C}$ be a length-$L$ multivariate sequence with $C$ channels, and $m\in\{0,1\}^{L}$ a binary mask, where $m_t=1$ indicates an observed time step and $m_t=0$ a missing one. Given the partially observed sequence and mask-aware features, the model generates missing values conditioned on the observed entries. We consider three mask schemes: (i) Random mask, which masks a random subset of time steps for interpolation/imputation, with missing fraction set by \texttt{ratio\_range}; (ii) Block mask, which masks a contiguous block at the end of the sequence for forecasting, with the prediction horizon controlled by \texttt{pred\_len\_range}; and (iii) Stride mask, which masks multiple blocks according to a structured stride pattern controlled by \texttt{num\_blocks\_range}, targeting interpolation/imputation under structured missingness. More details are provided in Appendix ~\ref{app:exp-details}.

\paragraph{Training and DP-SGD setup.}
All models are trained using identical settings: AdamW (learning rate $=7\times 10^{-4}$, weight decay $=2\times 10^{-5}$), $1000$-step linear warmup, mixed precision, and exponential moving average (EMA) with decay $0.999$. Training is run for $T=20{,}000$ steps ($200$ steps per epoch) with batch size $B=96$, and validation is performed every $1000$ steps. DP is enforced via DP-SGD with per-example gradient clipping at threshold $C=1.0$ and Gaussian noise injection calibrated by noise multiplier $\sigma$. DP-aware modifies only the forward pass via bounded conditioning, leaving DP-SGD unchanged.

\subsection{Model Utility under Matched DP Mechanisms}
\label{sec:main-results}

We compare non-private training (\textit{Non-DP}), vanilla DP-SGD (\textit{DP-vanilla}), and DP-SGD augmented with our proposed DP-Aware AdaLN-Zero (\textit{DP-aware}) on PrivatePower. Table~\ref{tab:main-results-extended} reports representative metrics for interpolation/imputation and forecasting. For interpolation, we assess point-wise accuracy via point\_RMSE, point\_MAPE, and point\_MAE. For forecasting, we report point\_RMSE talong with a distributional metric (dist\_JS) and a temporal-structure metric (temp\_spec\_dist).

Table \ref{tab:main-results-extended} shows that DP-aware consistently outperforms DP-vanilla across both tasks and all tested noise multipliers. Gains are largest in the low-noise regime, yet remain evident as noise increases. This corroborates our central claim. For fixed DP hyperparameters, bounding conditioning-induced gain improves conditional generation utility. Full per-$\sigma$ results are provided in Appendix ~\ref{app:extended-results}. 

 Figure~\ref{fig:grad_norm_distribution} further visualizes per-example gradient-norm distributions. DP-aware exhibits clear tail-suppression: while the bulk of its distribution closely aligns with DP-vanilla, extreme gradient events, particularly for $\|g_{\mathrm{cond}}\|_2$, occur less frequently. Together with Table~\ref{tab:main-results-extended}, this confirms that DP-aware reduces rare, condition-amplified gradient spikes that dominate global clipping, thus improving training stability and conditional generation utility.

\subsection{Gradient Dynamics and Clipping Behavior}
\label{sec:sensitivity}

Following Section \ref{sec:3.5}, we empirically study DP-SGD dynamics, starting with gradient-norm tails. During training, we record per-example gradient norms for conditioning-path parameters ($\|g_{\mathrm{cond}}\|_2$) and all other parameters ($\|g_{\mathrm{other}}\|_2$). For each run, we summarize these norms using robust high-percentile statistics ($p95$/$p99$) over all logged steps and samples, denoted as $S_{\text{cond}}^{(\cdot)}$ and $S_{\text{other}}^{(\cdot)}$, along with a compact proxy $S_{\text{total}}$. As average-case diagnostics, we also report $\rho_{\text{emp}}$ and $\rho_{\text{cond}}$, which are empirical summaries of observed scale/tail changes relative to DP-vanilla. Table~\ref{tab:grad-norms-multi-m} reveals a consistent trend across noise multipliers: DP-aware only mildly affects $S_{\text{other}}$, but markedly reduces the tail of $S_{\text{cond}}$, especially at $p99$. This demonstrates that by bounding the conditioning gain in the forward pass, DP-aware mainly suppresses rare condition-amplified outliers instead of uniformly shrinking gradients.

\begin{table}[t]
    \centering
    \resizebox{0.43\textwidth}{!}{
    \begin{tabular}{l l c c c c c}
        \toprule
        $\sigma$ & Model & $p_{\mathrm{clip}}$ & $\mathbb{E}[\eta]$ & $\eta_{10}$ & $\eta_{50}$ & $\eta_{90}$ \\
        \midrule
        \multirow{2}{*}{0.03}
            & DP-vanilla & 0.561   & 0.73   & 0.18   & 0.86   & 1.00   \\
            & DP-aware   & 0.556   & 0.75   & 0.22   & 0.88   & 1.00   \\
        \addlinespace[0.4em]
        \multirow{2}{*}{0.20}
            & DP-vanilla & 0.589   & 0.70   & 0.12   & 0.82   & 1.00   \\
            & DP-aware   & 0.585   & 0.72   & 0.15   & 0.84   & 1.00   \\
        \bottomrule
    \end{tabular}}
    \caption{\textbf{Clipping statistics under matched $C$.} We report the clipping rate $p_{\mathrm{clip}}$ and quantiles of the clipping factor
    $\eta$. Across noise multipliers, $p_{\mathrm{clip}}$ remains comparable, while DP-aware slightly increases low-quantile and median values of $\eta$, suggesting reduced clipping severity.}
    \label{tab:clipping_stats}
\end{table}
\vspace{-2mm}

 \begin{table*}[ht]
    \centering
    \resizebox{0.93\textwidth}{!}{
        \begin{tabular}{lccccc}
        \toprule
        Model Variant & $\mathbf{c}$ bounded & $(\gamma,\beta,\alpha)$ bounded & $\rho_{\text{emp}}$ &
        Forecast point\_RMSE $\downarrow$ &
        Forecast dist\_JS $\downarrow$ \\
        \midrule
        DP-vanilla & $\times$ & $\times$ & 1.00 & 0.567   & 0.723   \\
        DP-aware (only $\mathbf{c}$-bounding) & $\checkmark$ & $\times$ & 0.95 & 0.510   & 0.691   \\
        DP-aware (only AdaLN bounding) & $\times$ & $\checkmark$ & 0.93 & 0.495   & 0.677   \\
        DP-aware (full, ours) & $\checkmark$ & $\checkmark$ & 0.87 & \textbf{0.423}   & \textbf{0.636}   \\
        \bottomrule
        \end{tabular}}
    \caption{\textbf{Effects of forward-pass control components.} We report $\rho_{\text{emp}}(p99)$ and forecasting metrics (lower is better).}
    \label{tab:ablation-forecast}
\end{table*}

Tail behavior is crucial in DP-SGD, as large gradient norms trigger clipping and distort updates. For a clipping threshold $C$, we measure clipping activation rate $p_{\mathrm{clip}} \coloneqq \Pr\!\big[\|g_i\|_2 > C\big]$ and clipping severity $\eta_i \coloneqq\min\!\left(1,\frac{C}{\|g_i\|_2}\right)$, where smaller $\eta_i$ means stronger rescaling. Table~\ref{tab:clipping_stats} shows that under matched $(C,\sigma)$, DP-vanilla and DP-aware have comparable $p_{\mathrm{clip}}$, while DP-aware yields slightly larger $\eta$ at lower quantiles and around the median, implying milder clipping. Combined with the observed reduction in tail gradient norms in Table~\ref{tab:grad-norms-multi-m}, particularly along the conditioning pathway, this suggests that DP-aware preserves similar clipping frequency while attenuating extreme gradient spikes, hence fewer severe clipping events. More results are presented in Appendix~\ref{app:clip-norm}.

Overall, under matched parameter settings, DP-aware improves utility by altering optimization dynamics: DP-vanilla is more prone to conditioning-induced tail events that trigger global clipping and distort updates; whereas DP-aware suppresses such outliers in the forward pass, making conditional signals more robust to clipping. We provide more training-time signals in Appendix ~\ref{app:training-dynamics}.

\subsection{Ablation Studies}
\label{sec:ablations}

Next, we investigate the impact of key design choices in DP-aware AdaLN-Zero on downstream utility. We perform ablations on PrivatePower for forecasting, using identical DP hyperparameters for DP-vanilla and DP-aware variants. 

\subsubsection{Effects of Forward-Pass Control Components}
\label{sec:component_ablation}
DP-aware applies two forward-pass controls: $\ell_2$-norm bounding of conditioning vector $\mathbf{c}$ and coordinate-wise bounding of AdaLN modulations $(\gamma,\beta,\alpha)$. We ablate these components by comparing \emph{only $\mathbf{c}$-bounding}, \emph{only AdaLN bounding}, \emph{full DP-aware model} and the \emph{DP-vanilla} model. Table~\ref{tab:ablation-forecast} shows that each component provides a measurable gain, and their combination achieves the best utility, highlighting the importance of regulating both global conditioning and per-block modulation for stable conditional learning under DP-SGD.

\subsubsection{Effect of the AdaLN Modulation Bounding Operator}
\label{sec:boundop}

DP-aware bounds AdaLN modulation parameters $(\gamma,\beta,\alpha)$ via a coordinate-wise operator $\mathcal{B}_M(\cdot)$. Our default choice is $\mathcal{B}_M^{\tanh}(\cdot)$, but alternatives include hard truncation and near-identity bounds that deviate from identity only in a narrow boundary band. Concretely, we compare four realizations:
\begin{itemize}
    \item \texttt{tanh} (default): $\mathcal{B}_M^{\tanh}(x)=M\tanh(x/M)$;
    \item \texttt{hard\_clamp}: $\mathcal{B}_M^{\mathrm{hard}}(x)=\min(M,\max(-M,x))$;
    \item \texttt{soft\_clamp\_band}: identity for $|x|\le M-\varepsilon$, smooth transition in $[M-\varepsilon,\,M+\varepsilon]$, saturation for $|x|\ge M+\varepsilon$;
    \item \texttt{clamp\_ste}: the forward pass applies a hard clamp (\texttt{hard\_clamp}), while the backward pass uses a straight-through estimator.
\end{itemize}
Experimental results in Table~\ref{tab:boundop-mini} show that smooth operators (\texttt{tanh}, \texttt{soft\_clamp\_band}) perform comparably and consistently outperform hard truncation variants (\texttt{hard\_clamp}, \texttt{clamp\_ste}). All operators achieve better results than vanilla DP-SGD (point\_RMSE($0.567$), dist\_JS($0.643$), and temp\_spec\_dist(1.650e-3) ). This indicates that DP-aware’s gains mainly arise from bounding modulation to suppress extreme tail behavior, rather than from a particular function. Yet, in practice, DP-aware prefers smooth operators. See Appendix~\ref{app:C3} for more results across tasks and noise levels.

\begin{table}[ht]
    \centering
    \resizebox{0.48\textwidth}{!}{
    \begin{tabular}{lccc}
    \toprule
    $\mathcal{B}_M(\cdot)$
    & point\_RMSE$\downarrow$
    & dist\_JS$\downarrow$
    & temp\_spec\_dist$\downarrow$ \\
    \midrule
    $\tanh$ (default)           & 0.423 & 0.636 & 8.558e-4 \\
    \texttt{soft\_clamp\_band}  & 0.425 & 0.639 & 8.70e-4  \\
    \texttt{hard\_clamp}        & 0.452 & 0.662 & 9.45e-4  \\
    \texttt{clamp\_ste}         & 0.467 & 0.674 & 9.90e-4  \\
    \bottomrule
    \end{tabular}}
    \caption{\textbf{Effect of the bounding operator $\mathcal{B}_M(\cdot)$ at $\sigma{=}0.05$.}}
\label{tab:boundop-mini}
\end{table}

Appendix ~\ref{app:ablations} presents more ablations, examining the effects of tightness settings, DP-aware architectural constraints, clipping thresholds, and interpolation/imputation metrics.

\section{Conclusion and Future Work}

Under DP-SGD, conditional diffusion models suffer a failure mode wherein conditioning amplifies sensitivity, producing heavy-tailed gradients. These outliers dominate the global clipping of per-example gradients, thus uniformly shrinking updates and inducing a systematic optimization bias that diffusion-specific DP improvements at the global mechanism level cannot resolve. To address this issue, we propose the DP-aware AdaLN-Zero, a sensitivity-aware conditioning module without modifying the DP-SGD mechanism. By jointly constraining the conditioning representation and the AdaLN modulation parameters, our approach selectively suppresses extreme gradient tails along the conditioning-path parameters, leading to milder clipping distortion.


Future work can extend our approach in several directions. We will integrate DP-aware conditioning with advanced DP diffusion pipelines, such as pretraining followed by DP fine-tuning, to further improve the privacy--utility trade-off. We will also replace the offline choices of $(c_{\max},\gamma_{\max},\beta_{\max},\alpha_{\max})$ with principled, privacy-preserving calibration methods. Layer- or block-wise adaptive bounds may improve robustness across datasets and tasks. In addition, we will extend the sensitivity-control principle to richer conditioning interfaces, including cross-attention, encoder--decoder conditioning, and multi-source context, to go beyond AdaLN-style modulation.



\bibliography{example_paper}
\bibliographystyle{icml2026}

\newpage
\appendix
\onecolumn
\section{Proof Details for Proposition 1}\label{appendix:proof_dp_aware}

In this section, we provide detailed derivations for Proposition 1 in the main text. We first state the assumptions used throughout the analysis. We then bound the Jacobians of a single AdaLN-Zero block, derive per-sample gradient bounds for each parameter group, and conclude by summarizing the resulting DP-SGD sensitivity bounds.

\subsection{Assumptions}
\label{appendix:assumptions}

We adopt the following assumptions, which are standard in worst-case sensitivity analysis.

\begin{assumption}[Bounded inputs]\label{assumption1}
    There exists a constant $X_{\max} > 0$ such that, for all training examples $x$,
    \begin{equation}
        \|x\|_2 \le X_{\max}.
    \end{equation} 
\end{assumption}

\begin{assumption}[LayerNorm boundedness and Lipschitzness]\label{assumption2}
    There exist constants $U_{\max} > 0$ and $L_{\mathrm{LN}} > 0$ such that, for all $x, x_1, x_2$,
    \begin{equation}
    \begin{aligned}
        \|\mathrm{LN}(x)\|_2 &\le U_{\max},
        \\
        \|\mathrm{LN}(x_1) - \mathrm{LN}(x_2)\|_2
        &\le
        L_{\mathrm{LN}} \|x_1 - x_2\|_2.
    \end{aligned}
    \end{equation}
\end{assumption}

\begin{assumption}[Feedforward/Attention boundedness and Lipschitzness]\label{assumption3}
    Let $F$ denote the feedforward or attention sub-layer. There exist constants $H_{\max} > 0$ and $L_F > 0$ such that, for all $v, v_1, v_2$ in the reachable domain of $F$,
    \begin{equation}
    \begin{aligned}
        \|F(v)\|_2 &\le H_{\max},
        \\
        \|F(v_1) - F(v_2)\|_2 &\le L_F \|v_1 - v_2\|_2.
    \end{aligned}
    \end{equation}
\end{assumption}

\begin{assumption}[Bounded loss gradients]\label{assumption4}
    There exists a constant $G_\ell > 0$ such that, for all reachable outputs $y$,
    \begin{equation}
        \|\nabla_y \ell(y)\|_2 \le G_\ell.
    \end{equation}
\end{assumption}

\begin{assumption}[Spectral norm bounds]\label{assumption5}
    Each linear layer $W$ in the network satisfies
    \begin{equation}
        \|W\|_{\mathrm{op}} \le \omega_{\max},
    \end{equation}
    and the architecture (e.g., the number of blocks and hidden dimensions) is finite.
\end{assumption}

Moreover, the DP-aware constraints ensure boundedness along the conditioning pathway. In particular, for each block,
\begin{equation}
\begin{aligned}
\|\mathbf{c}\|_2 \le c_{\max},\,&
    \|\gamma\|_\infty \le \gamma_{\max},\,\\
    \|\beta\|_\infty \le \beta_{\max},\,&
    \|\alpha\|_\infty \le \alpha_{\max}.
    \label{eq:dp_aware_bounds_appendix}
\end{aligned}
\end{equation}
Combined with Assumptions \ref{assumption1}-\ref{assumption5}, we bound intermediate Jacobians and gradient norms.

\subsection{Jacobian Bounds for an AdaLN-Zero Block}
\label{appendix:block_jacobian}

We consider a single AdaLN-Zero block defined as
\begin{equation}
    u = \mathrm{LN}(x), \,
    v = \gamma \odot u + \beta, \,
    h = F(v), \,
    y = x + \alpha \odot h.
\end{equation}
We follow this sequence of computations to derive Jacobian bounds for the AdaLN-Zero block.

\paragraph{From $u$ to $v$.}
The Jacobian of $v$ with respect to $u$ is
\begin{equation}
    \frac{\partial v}{\partial u}
    =
    \mathrm{diag}(\gamma),
\end{equation}
and therefore
\begin{equation}
    \Bigl\|\frac{\partial v}{\partial u}\Bigr\|_{\mathrm{op}}
    =
    \|\gamma\|_\infty
    \le
    \gamma_{\max}.
\end{equation}

\paragraph{From $v$ to $h$.}
By Assumption \ref{assumption3},
\begin{equation}
    \Bigl\|\frac{\partial h}{\partial v}\Bigr\|_{\mathrm{op}}
    \le L_F.
\end{equation}
Applying the chain rule yields
\begin{equation}
    \Bigl\|\frac{\partial h}{\partial u}\Bigr\|_{\mathrm{op}}
    \le
    L_F \gamma_{\max}.
\end{equation}

\paragraph{From $x$ to $y$.}
We have
\begin{equation}
    \frac{\partial y}{\partial x}
    =
    I
    +
    \mathrm{diag}(\alpha)
    \cdot
    \frac{\partial h}{\partial x}.
\end{equation}
Moreover, by the chain rule,
\begin{equation}
    \frac{\partial h}{\partial x}
    =
    \frac{\partial h}{\partial u}
    \cdot
    \frac{\partial u}{\partial x},
\end{equation}
and by Assumption \ref{assumption2},
\begin{equation}
    \Bigl\|\frac{\partial u}{\partial x}\Bigr\|_{\mathrm{op}}
    \le
    L_{\mathrm{LN}}.
\end{equation}

Combining the above bounds and using $\|\alpha\|_\infty \le \alpha_{\max}$, we obtain
\begin{equation}
    \Bigl\|\frac{\partial y}{\partial x}\Bigr\|_{\mathrm{op}}
    \le
    1
    +
    \alpha_{\max}\,L_F\,\gamma_{\max}\,L_{\mathrm{LN}}
    \;\eqqcolon\;
    C_{\mathrm{block}},
\end{equation}
where $C_{\mathrm{block}}$ depends only on architecture-dependent constants and DP-aware bounds. Applying the same argument across all blocks yields a global bound on the Jacobian of the full network with respect to its inputs and intermediate states.

\subsection{Gradient Bounds by Parameter Groups}
\label{appendix:gradient_groups}

Let $\theta$ denote the collection of all model parameters. For a single example $z=(x,\mathbf{c})$, define the per-sample gradient as
\begin{equation}
    g_z(\theta) \;=\; \nabla_\theta \ell\!\big(f_\theta(x,\mathbf{c})\big).
\end{equation}
We partition $\theta$ into three parameter groups and bound the contribution of each group separately.

\subsubsection*{Group 1: Parameters in $F$}

Consider a weight matrix $W$ in $F$. Its corresponding gradient can be expressed as
\begin{equation}
    \nabla_W \ell
    =
    (\alpha \odot \nabla_y \ell)
    \cdot
    \phi(v)^\top,
\end{equation}
where $\phi(v)$ denotes the feature vector derived from $v$. By Assumption~\ref{assumption4} and Eq.(\ref{eq:dp_aware_bounds_appendix}), we obtain
\begin{align}
    \|\alpha \odot \nabla_y \ell\|_2
    &\le
    \|\alpha\|_\infty \|\nabla_y \ell\|_2
    \le
    \alpha_{\max} G_\ell, \\
    \|\phi(v)\|_2
    &\le
    H_{\max}
    +
    \gamma_{\max} U_{\max}
    +
    \sqrt{d}\,\beta_{\max},
\end{align}
where $d$ is the hidden dimension. Hence,
\begin{equation}
    \|\nabla_W \ell\|_F
    \le
    a_{\alpha,F}\,\alpha_{\max}
    +
    a_{\gamma,F}\,\gamma_{\max}
    +
    a_{\beta,F}\,\beta_{\max},
\end{equation}
for suitable non-negative constants $a_{\alpha,F}, a_{\gamma,F},a_{\beta,F}$. Summing the above bound over all weight matrices $W\in F$ yields an aggregate contribution of the form: $a_{\alpha}\alpha_{\max}+a_{\gamma}\,\gamma_{\max}+a_{\beta}\,\beta_{\max}$, where $a_{\alpha}, a_{\gamma}, a_{\beta} \ge 0$ collect the corresponding per-layer constants and depend only on the model architecture. Equivalently, the sensitivity contribution from the parameters within $F$ scales linearly in $(\alpha_{\max}, \gamma_{\max}, \beta_{\max})$.

\subsubsection*{Group 2: Projection layers for $(\gamma,\beta,\alpha)$}

The modulation parameters are obtained from $\mathbf{c}$ via linear projections followed by clipping,
\begin{equation}
    \gamma_{\mathrm{raw}} = W_\gamma \mathbf{c} + b_\gamma,
    \,
    \gamma = \mathrm{clip}(\gamma_{\mathrm{raw}},\gamma_{\max}).
\end{equation}
Because coordinate-wise clipping is $1$-Lipschitz, it cannot amplify backpropagated gradients; in particular,
$\left\|\frac{\partial \ell}{\partial \gamma_{\mathrm{raw}}}\right\|_2 \le \bigl\|\frac{\partial \ell}{\partial \gamma}\bigr\|_2$. Under the boundedness assumptions in Group~1, the upstream gradient is therefore bounded in norm. Hence,
\begin{equation}
    \nabla_{W_\gamma} \ell
    =
    \left(\frac{\partial \ell}{\partial \gamma_{\mathrm{raw}}} \right)\,\mathbf{c}^\top.
\end{equation}
Using $\|\mathbf{c}\|_2 \le c_{\max}$, we obtain
\begin{equation}
    \|\nabla_{W_\gamma} \ell\|_F
    \le
    c_\gamma\,c_{\max}
\end{equation}
for some constant $c_\gamma \ge 0$. Analogous bounds hold for $W_\beta$ and $W_\alpha$. Summing over these projection layers yields a contribution of the form $a_c\,c_{\max}$, where $a_c \ge 0$ is an architecture-dependent constant.

\subsubsection*{Group 3: Remaining parameters}

All remaining parameters (e.g., LayerNorm weights, skip connections) depend only on bounded activations and receive bounded upstream gradients. Consequently, their contribution can be bounded by a constant $A_0 \ge 0$ that is independent of $(c_{\max},\gamma_{\max},\beta_{\max},\alpha_{\max})$.

\subsection{Combining Per-Sample Gradient Bounds}
\label{appendix:combining_bounds}

Combining the bounds from the three groups yields
\begin{equation}
    \bigl\| g_z(\theta) \bigr\|_2
    \;\le\;
    A_0
    + a_c c_{\max}
    + a_\gamma \gamma_{\max}
    + a_\beta \beta_{\max}
    + a_\alpha \alpha_{\max}.
\end{equation}
This matches Proposition 1, where the per-sample gradient bound under the DP-aware constraints is
\begin{equation}
    S_{\mathrm{aware}}
    =
    A_0
    + a_c c_{\max}
    + a_\gamma \gamma_{\max}
    + a_\beta \beta_{\max}
    + a_\alpha \alpha_{\max}.
\end{equation}

\section{DP-SGD Sensitivity Comparison}
\label{appendix:sensitivity_details}

In this section, we compare the $\ell_2$-sensitivity of vanilla DP-SGD and our DP-aware variant. We further introduce a normalized sensitivity ratio, $\tfrac{S_{\mathrm{aware}}}{S_{\mathrm{vanilla}}^{\mathrm{ref}}}$, and show that it admits a convex-combination representation.

\paragraph{Vanilla DP-SGD.}
Given per-sample gradients $\{g_i\}_{i=1}^B$ and a clipping threshold $C>0$, vanilla DP-SGD applies per-sample gradient clipping:
\begin{equation}
    \tilde g_i
    =
    g_i \cdot \min\!\left(1, \frac{C}{\|g_i\|_2}\right),
    \,
    q(D) = \frac{1}{B} \sum_{i=1}^B \tilde g_i.
\end{equation}
By construction, $\|\tilde g_i\|_2 \le C$ for all $i$. Therefore, for any neighboring batches $D,D'$ that differ in at most one example, the $\ell_2$-sensitivity satisfies
\begin{equation}
    \Delta_2(q_{\mathrm{vanilla}})
    \le \frac{C}{B}.
\end{equation}

\paragraph{DP-aware DP-SGD.}
For the DP-aware model, assume the per-sample gradients are uniformly bounded as $\|g_i\|_2 \le S_{\mathrm{aware}}$ for all $i$. If $S_{\mathrm{aware}} \le C$, clipping is inactive and hence
\begin{equation}
    \Delta_2(q_{\mathrm{aware}})
    \le \frac{S_{\mathrm{aware}}}{B}.
\end{equation}
Consequently, the sensitivity ratio
\begin{equation}
    \rho
    =
    \frac{\Delta_2(q_{\mathrm{aware}})}{\Delta_2(q_{\mathrm{vanilla}})}
    \le
    \frac{S_{\mathrm{aware}}}{C}.
\end{equation}

\paragraph{Normalized ratio and convex combination form.}
To relate $S_{\mathrm{aware}}$ to a representative scale in a standard (non-DP) training run, we introduce non-negative reference magnitudes
\begin{equation}
    C_{\mathrm{ref}},\,
    \Gamma_{\mathrm{ref}},\,
    B_{\mathrm{ref}},\,
    A_{\mathrm{ref}},
\end{equation}
which summarize typical magnitudes of $\|\mathbf{c}\|_2$, $|\gamma|$, $|\beta|$, and $|\alpha|$, respectively. Using the same architectural constants, define the corresponding reference sensitivity scale
\begin{equation}
    S_{\mathrm{vanilla}}^{\mathrm{ref}}
    =
    A_0
    + a_c C_{\mathrm{ref}}
    + a_\gamma \Gamma_{\mathrm{ref}}
    + a_\beta B_{\mathrm{ref}}
    + a_\alpha A_{\mathrm{ref}}.
\end{equation}

Let
\begin{align}
    u_0 = A_0,\,&u_c = a_c C_{\mathrm{ref}},\\u_\gamma = a_\gamma \Gamma_{\mathrm{ref}},\,u_\beta = &a_\beta B_{\mathrm{ref}},\,u_\alpha = a_\alpha A_{\mathrm{ref}},
\end{align}
and denote their sum by
\begin{equation}
    D
    =
    u_0 + u_c + u_\gamma + u_\beta + u_\alpha
    =
    S_{\mathrm{vanilla}}^{\mathrm{ref}}.
\end{equation}

Define weights
\begin{equation}
    \lambda_0 = \frac{u_0}{D},\,
    \lambda_c = \frac{u_c}{D},\,
    \lambda_\gamma = \frac{u_\gamma}{D},\,
    \lambda_\beta = \frac{u_\beta}{D},\,
    \lambda_\alpha = \frac{u_\alpha}{D},
\end{equation}
which satisfy $\lambda_\cdot \ge 0$ and sum to $1$, and define the relative factors
\begin{equation}
    r_c = \frac{c_{\max}}{C_{\mathrm{ref}}},\,
    r_\gamma = \frac{\gamma_{\max}}{\Gamma_{\mathrm{ref}}},\,
    r_\beta = \frac{\beta_{\max}}{B_{\mathrm{ref}}},\,
    r_\alpha = \frac{\alpha_{\max}}{A_{\mathrm{ref}}}.
\end{equation}

We can then rewrite
{\small
\begin{align}
    S_{\mathrm{aware}}
    &=
    A_0 + a_c c_{\max} + a_\gamma \gamma_{\max} + a_\beta \beta_{\max} + a_\alpha \alpha_{\max} \\
    &=
    u_0
    + u_c r_c
    + u_\gamma r_\gamma
    + u_\beta r_\beta
    + u_\alpha r_\alpha.
\end{align}}

Dividing both sides by $S_{\mathrm{vanilla}}^{\mathrm{ref}} = D$ yields the convex-combination form:
\begin{equation}
    \frac{S_{\mathrm{aware}}}{S_{\mathrm{vanilla}}^{\mathrm{ref}}}
    =
    \lambda_0
    + \lambda_c r_c
    + \lambda_\gamma r_\gamma
    + \lambda_\beta r_\beta
    + \lambda_\alpha r_\alpha.
\end{equation}
In particular,
\begin{equation}
    \frac{S_{\mathrm{aware}}}{S_{\mathrm{vanilla}}^{\mathrm{ref}}}
    \le
    \max\{1, r_c, r_\gamma, r_\beta, r_\alpha\}.
\end{equation}
If
\begin{equation}
\begin{aligned}
    c_{\max} \le C_{\mathrm{ref}},\,
    \gamma_{\max} \le \Gamma_{\mathrm{ref}},\,
    \beta_{\max} \le B_{\mathrm{ref}},\,
    \alpha_{\max} \le A_{\mathrm{ref}},
\end{aligned}
\end{equation}
then $r_c,r_\gamma,r_\beta,r_\alpha \le 1$, implying
\begin{equation}
    \frac{S_{\mathrm{aware}}}{S_{\mathrm{vanilla}}^{\mathrm{ref}}}
    \le 1.
\end{equation}
Combined with $\rho \le \frac{S_{\mathrm{aware}}}{C}$, this yields $\rho \le 1$ whenever $C$ is chosen to be at least the corresponding reference scale.

\section{Additional Results on Model Utility}
\label{app:extended-results}

\subsection{Per-$\sigma$ Results on PrivatePower}\label{app:full-metrics}

\begin{table*}[t]
    \centering
    \resizebox{0.94\textwidth}{!}{%
        \begin{tabular}{lcccccccccc}
            \toprule
            Model &
            $\sigma$ &
            \begin{tabular}[c]{@{}c@{}} MAE $\downarrow$ \end{tabular} &
            \begin{tabular}[c]{@{}c@{}} RMSE $\downarrow$ \end{tabular} &
            \begin{tabular}[c]{@{}c@{}} MAPE $\downarrow$ \end{tabular} &
            \begin{tabular}[c]{@{}c@{}} R2 $\uparrow$ \end{tabular} &
            \begin{tabular}[c]{@{}c@{}} dist\_KL $\downarrow$ \end{tabular} &
            \begin{tabular}[c]{@{}c@{}} dist\_JS $\downarrow$ \end{tabular} &
            \begin{tabular}[c]{@{}c@{}} dist\_WS $\downarrow$ \end{tabular} &
            \begin{tabular}[c]{@{}c@{}} dist\_KS $\downarrow$ \end{tabular} &
            \begin{tabular}[c]{@{}c@{}} MMD $\downarrow$ \end{tabular} \\
            \midrule

            Non-DP (no DP) &
            -- &
            0.155 & 0.208 & 1.90\% & 0.861 &
            0.980 & 0.767 & 0.605 & 0.215 & 1.48e-2 \\
            \midrule

            DP-vanilla &
            0.03 &
            0.430 & 0.556 & 4.85\% & 0.422 &
            1.122 & 0.736 & 0.820 & 0.260 & 3.90e-2 \\
            DP-aware &
            0.03 &
            \textbf{0.235} & \textbf{0.296} & \textbf{3.10\%} &
            \textbf{0.741} &
            \textbf{1.025} & \textbf{0.684} & \textbf{0.700} &
            \textbf{0.230} & \textbf{2.70e-2} \\
            \midrule

            DP-vanilla &
            0.05 &
            0.448 & 0.567 & 5.10\% & 0.383 &
            1.081 & 0.643 & 0.840 & 0.270 & 4.80e-2 \\
            DP-aware &
            0.05 &
            \textbf{0.325} & \textbf{0.423} & \textbf{4.00\%} &
            \textbf{0.603} &
            \textbf{1.052} & \textbf{0.636} & \textbf{0.810} &
            \textbf{0.258} & \textbf{4.10e-2} \\
            \midrule

            DP-vanilla &
            0.1 &
            1.280 & 1.637 & 12.9\% & -0.105 &
            1.555 & 0.833 & 1.458 & 0.390 & 1.90e-1 \\
            DP-aware &
            0.1 &
            \textbf{0.520} & \textbf{0.671} & \textbf{6.10\%} &
            \textbf{0.353} &
            \textbf{1.256} & \textbf{0.757} & \textbf{0.982} &
            \textbf{0.310} & \textbf{9.00e-2} \\
            \midrule

            DP-vanilla &
            0.2 &
            1.310 & 1.646 & 13.4\% & -0.122 &
            1.480 & 0.794 & 1.386 & 0.370 & 1.70e-1 \\
            DP-aware &
            0.2 &
            \textbf{1.000} & \textbf{1.262} & \textbf{10.1\%} &
            \textbf{0.052} &
            \textbf{1.347} & \textbf{0.732} & \textbf{1.205} &
            \textbf{0.330} & \textbf{1.35e-1} \\
            \bottomrule
        \end{tabular}%
    }
    \caption{\textbf{Forecasting metrics on PrivatePower across noise multipliers $\sigma$.} Lower is better for all metrics except R2.}
    \label{tab:forecasting-full}
\end{table*}

\begin{table*}[t]
    \centering
    \resizebox{0.94\textwidth}{!}{%
        \begin{tabular}{lcccccccccc}
            \toprule
            Model &
            $\sigma$ &
            \begin{tabular}[c]{@{}c@{}} MAE $\downarrow$ \end{tabular} &
            \begin{tabular}[c]{@{}c@{}} RMSE $\downarrow$ \end{tabular} &
            \begin{tabular}[c]{@{}c@{}} MAPE $\downarrow$ \end{tabular} &
            \begin{tabular}[c]{@{}c@{}} R2 $\uparrow$ \end{tabular} &
            \begin{tabular}[c]{@{}c@{}} dist\_KL $\downarrow$ \end{tabular} &
            \begin{tabular}[c]{@{}c@{}} dist\_JS $\downarrow$ \end{tabular} &
            \begin{tabular}[c]{@{}c@{}} dist\_WS $\downarrow$ \end{tabular} &
            \begin{tabular}[c]{@{}c@{}} dist\_KS $\downarrow$ \end{tabular} &
            \begin{tabular}[c]{@{}c@{}} MMD $\downarrow$ \end{tabular} \\
            \midrule

            Non-DP (no DP) &
            -- &
            0.476 & 0.584 & 1.17\% & 0.775 &
            0.900 & 0.660 & 0.560 & 0.190 & 1.20e-2 \\
            \midrule

            DP-vanilla &
            0.03 &
            2.391 & 3.034 & 5.67\% & -0.950 &
            1.210 & 0.820 & 0.760 & 0.240 & 3.20e-2 \\
            DP-aware &
            0.03 &
            \textbf{1.411} & \textbf{1.804} & \textbf{3.54\%} & \textbf{-0.250} &
            \textbf{1.030} & \textbf{0.740} & \textbf{0.670} & \textbf{0.220} & \textbf{2.40e-2} \\
            \midrule

            DP-vanilla &
            0.05 &
            3.110 & 3.498 & 9.34\% & -1.400 &
            1.260 & 0.835 & 0.790 & 0.250 & 3.80e-2 \\
            DP-aware &
            0.05 &
            \textbf{1.987} & \textbf{2.019} & \textbf{4.97\%} & \textbf{-0.600} &
            \textbf{1.090} & \textbf{0.760} & \textbf{0.700} & \textbf{0.230} & \textbf{2.90e-2} \\
            \midrule

            DP-vanilla &
            0.1 &
            4.702 & 5.787 & 11.67\% & -4.100 &
            1.520 & 0.910 & 1.080 & 0.310 & 9.50e-2 \\
            DP-aware &
            0.1 &
            \textbf{2.122} & \textbf{2.718} & \textbf{5.27\%} & \textbf{-1.100} &
            \textbf{1.280} & \textbf{0.820} & \textbf{0.860} & \textbf{0.260} & \textbf{5.80e-2} \\
            \midrule

            DP-vanilla &
            0.2 &
            5.148 & 6.812 & 12.65\% & -5.200 &
            1.600 & 0.930 & 1.250 & 0.340 & 1.20e-1 \\
            DP-aware &
            0.2 &
            \textbf{3.442} & \textbf{4.689} & \textbf{8.41\%} & \textbf{-2.200} &
            \textbf{1.430} & \textbf{0.870} & \textbf{1.050} & \textbf{0.300} & \textbf{8.50e-2} \\
            \bottomrule
        \end{tabular}%
    }
    \caption{\textbf{Interpolation/imputation metrics on PrivatePower across noise multipliers $\sigma$.}
    Lower is better for all metrics except R2.}
    \label{tab:inter-full}
\end{table*}

\begin{figure*}[t]
    \centering
    \subfigure[\textbf{ECDF} of gradient norms at noise multiplier $\sigma=0.03$]{
        \includegraphics[width=0.485\textwidth]{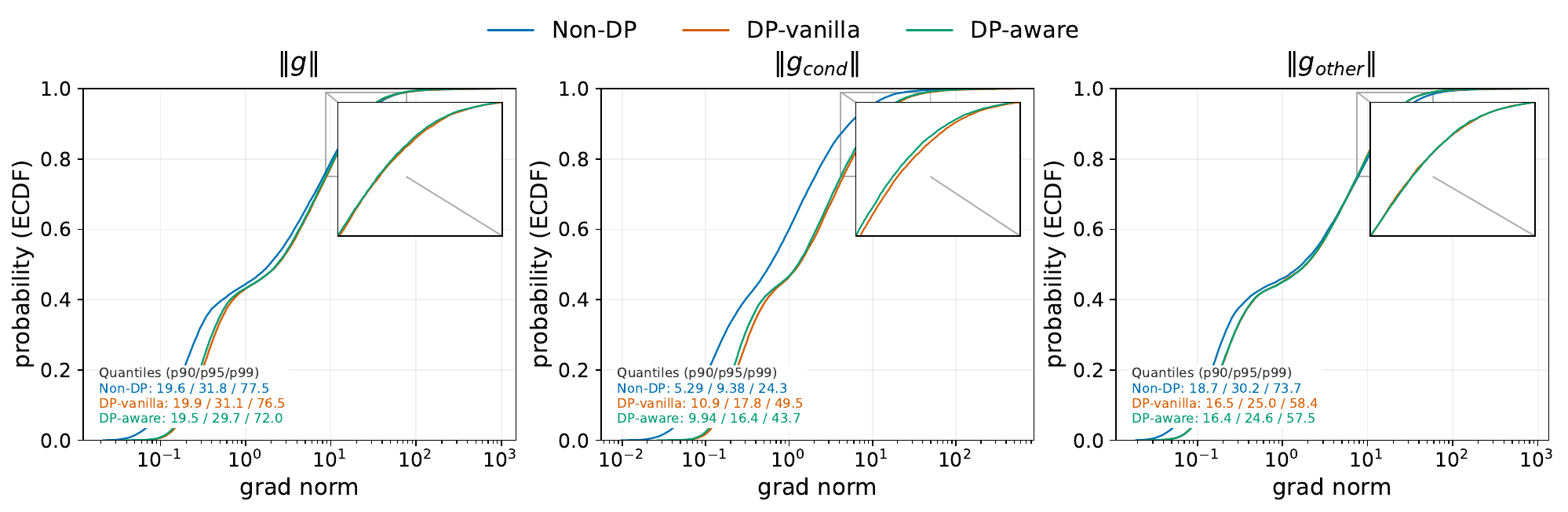}
    }
    \subfigure[\textbf{CCDF (tail)} of gradient norms at noise multiplier $\sigma=0.03$]{
        \includegraphics[width=0.485\textwidth]{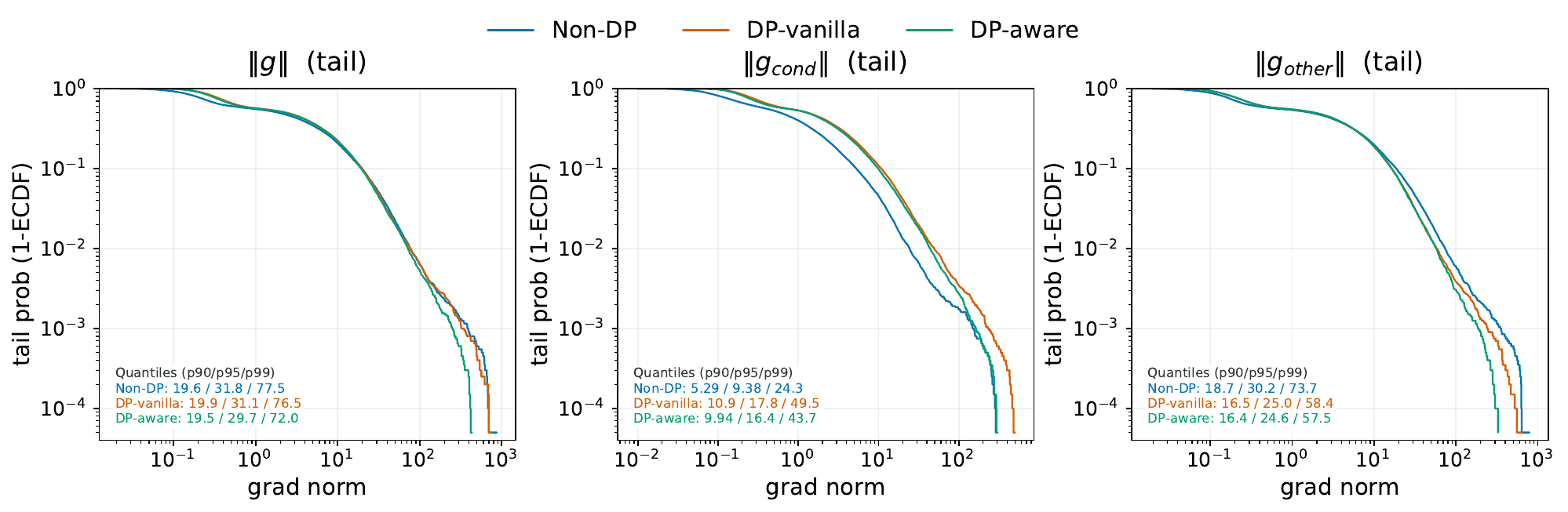}
    }
    \subfigure[\textbf{ECDF} of gradient norms at noise multiplier $\sigma=0.05$]{
        \includegraphics[width=0.485\textwidth]{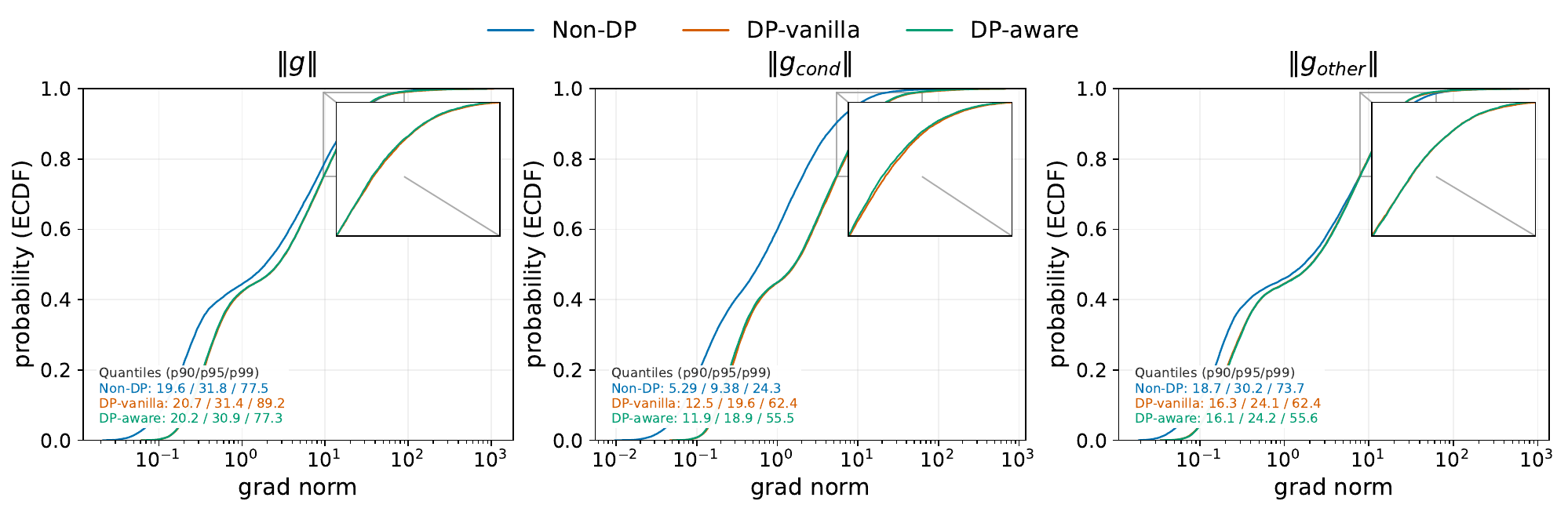}
        \label{ecdf:noise_multiplier=0.05}
    }
    \subfigure[\textbf{CCDF (tail)} of gradient norms at noise multiplier $\sigma=0.05$]{
        \includegraphics[width=0.485\textwidth]{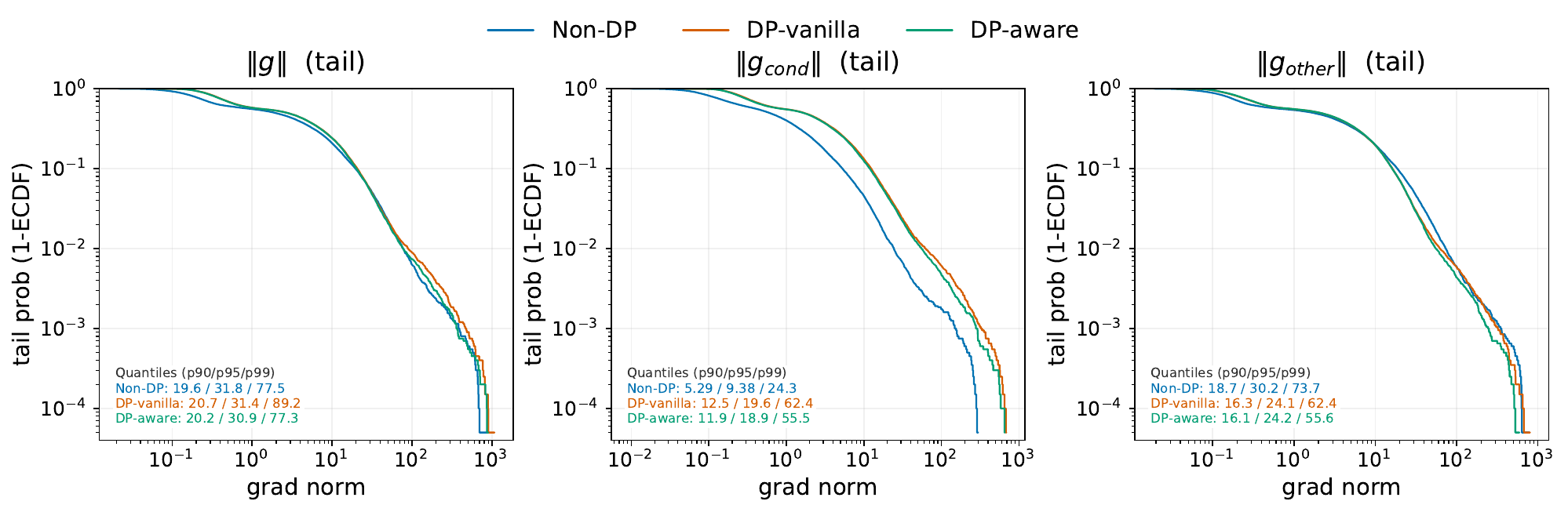}
        \label{tail:noise_multiplier=0.05}
    }
    \subfigure[\textbf{ECDF} of gradient norms at noise multiplier $\sigma=0.10$]{
        \includegraphics[width=0.485\textwidth]{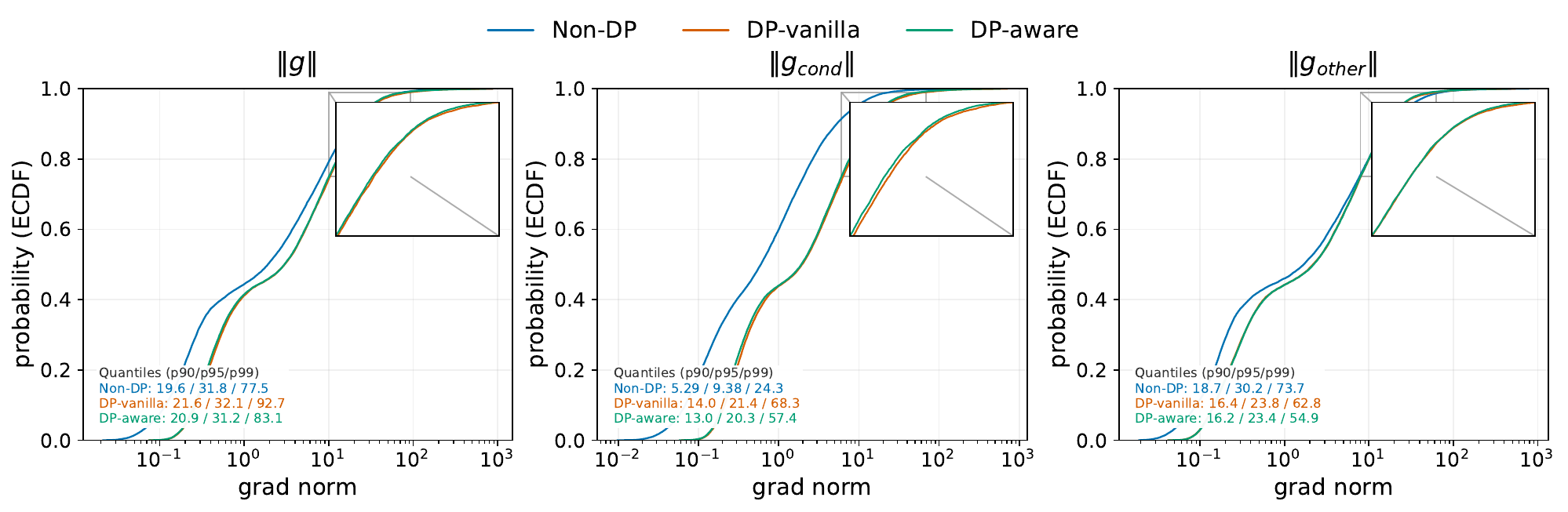}
        \label{ecdf:noise_multiplier=0.10}
    }
    \subfigure[\textbf{CCDF (tail)} of gradient norms at noise multiplier $\sigma=0.10$]{
        \includegraphics[width=0.485\textwidth]{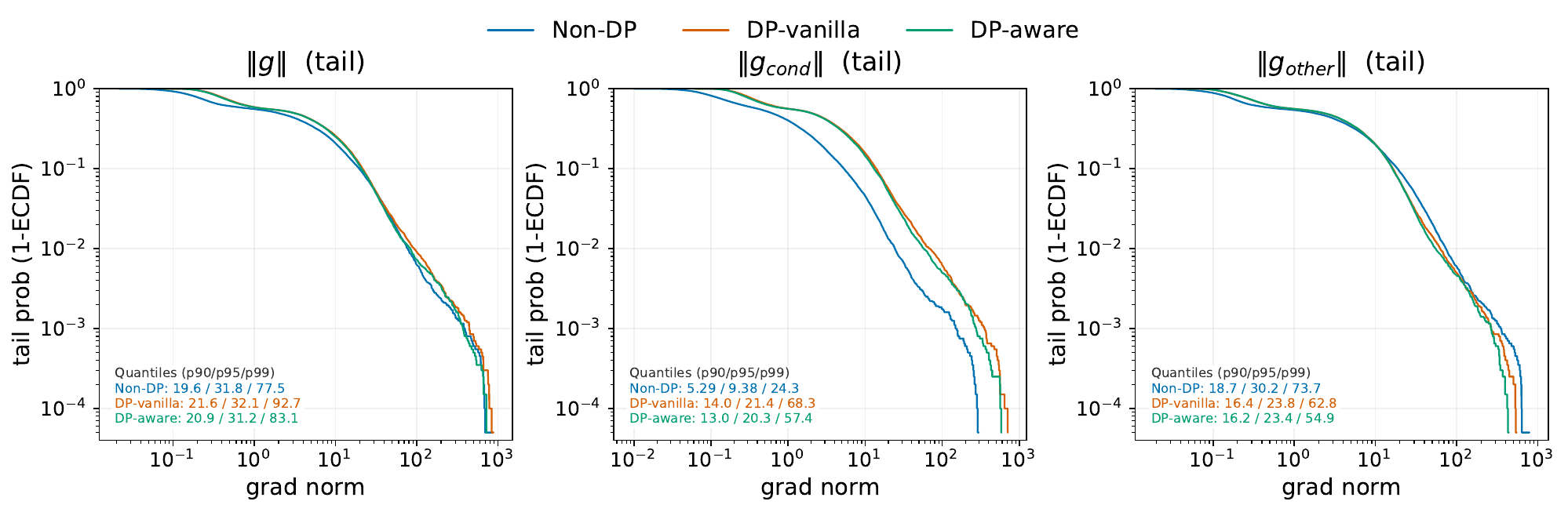}
        \label{tail:noise_multiplier=0.10}
    }
    \subfigure[\textbf{ECDF} of gradient norms at noise multiplier $\sigma=0.20$]{
        \includegraphics[width=0.485\textwidth]{images/hourly/fig_gradnorm_0.02_ecdf.pdf}
    }
    \subfigure[\textbf{CCDF (tail)} of gradient norms at noise multiplier $\sigma=0.20$]{
        \includegraphics[width=0.485\textwidth]{images/hourly/fig_gradnorm_0.02_ccdf_tail.pdf}
    }
    \caption{\textbf{Gradient distributions under DP training (all $\sigma$ settings).}}
    \label{fig:grad_norm_distribution_extented}
\end{figure*}

This section reports the complete set of metrics of our evaluation scripts on PrivatePower, complementing the representative results shown in the main text. Tables~\ref{tab:forecasting-full} and~\ref{tab:inter-full} summarize forecasting and interpolation/imputation performance, respectively. Figure~\ref{fig:grad_norm_distribution_extented} shows the per-example gradient-norm distributions observed during DP-SGD training.

\subsection{Results on ETTh1 and ETTm1}

\begin{table*}[!ht]
    \centering
    \resizebox{0.92\textwidth}{!}{
    \begin{tabular}{lccc|ccc}
        \toprule
        Model &
        \multicolumn{3}{c}{Interpolation/Imputation} &
        \multicolumn{3}{c}{Forecasting} \\
        \cmidrule(lr){2-4}\cmidrule(lr){5-7}
        &
        \begin{tabular}[c]{@{}c@{}}
        point\_RMSE $\downarrow$
        \end{tabular} &
        \begin{tabular}[c]{@{}c@{}}
        point\_MAPE $\downarrow$
        \end{tabular} &
        \begin{tabular}[c]{@{}c@{}}
        point\_MAE $\downarrow$
        \end{tabular} &
        \begin{tabular}[c]{@{}c@{}}
        point\_RMSE $\downarrow$
        \end{tabular} &
        \begin{tabular}[c]{@{}c@{}}
        dist\_JS $\downarrow$
        \end{tabular} &
        \begin{tabular}[c]{@{}c@{}}
        temp\_spectral\_dist $\downarrow$
        \end{tabular} \\
        \midrule
        Non-DP (no DP) & 0.310 & 2.35\% & 0.210 & 0.245 & 0.610 & 2.10e-4 \\
        DP-vanilla     & 0.780 & 5.90\% & 0.520 & 0.655 & 0.880 & 1.85e-3 \\
        DP-aware       & \textbf{0.560} & \textbf{4.10\%} & \textbf{0.380} &
                        \textbf{0.465} & \textbf{0.780} & \textbf{9.60e-4} \\
        \bottomrule
    \end{tabular}}
    \caption{\textbf{Representative metrics on ETTh1 (matched DP mechanism).}
    DP-vanilla and DP-aware are trained under matched DP-SGD settings ($C=1.0$, noise multiplier $\sigma = 0.1$, batch size $B=96$, and steps $T=20,000$).}
    \label{tab:main-results-bts}
\end{table*}

\begin{table*}[!ht]
    \centering
    \resizebox{0.92\textwidth}{!}{
    \begin{tabular}{lccc|ccc}
        \toprule
        Model &
        \multicolumn{3}{c}{Interpolation/Imputation} &
        \multicolumn{3}{c}{Forecasting} \\
        \cmidrule(lr){2-4}\cmidrule(lr){5-7}
        &
        \begin{tabular}[c]{@{}c@{}}
        point\_RMSE $\downarrow$
        \end{tabular} &
        \begin{tabular}[c]{@{}c@{}}
        point\_MAPE $\downarrow$
        \end{tabular} &
        \begin{tabular}[c]{@{}c@{}}
        point\_MAE $\downarrow$
        \end{tabular} &
        \begin{tabular}[c]{@{}c@{}}
        point\_RMSE $\downarrow$
        \end{tabular} &
        \begin{tabular}[c]{@{}c@{}}
        dist\_WS $\downarrow$
        \end{tabular} &
        \begin{tabular}[c]{@{}c@{}}
        temp\_spectral\_dist $\downarrow$
        \end{tabular} \\
        \midrule
        Non-DP (no DP) & 0.355 & 2.80\% & 0.240 & 0.295 & 0.640 & 2.60e-4 \\
        DP-vanilla     & 0.945 & 6.80\% & 0.625 & 0.805 & 0.920 & 2.15e-3 \\
        DP-aware       & \textbf{0.685} & \textbf{4.90\%} & \textbf{0.455} &
                        \textbf{0.565} & \textbf{0.820} & \textbf{1.20e-3} \\
        \bottomrule
    \end{tabular}}
    \caption{\textbf{Representative metrics on ETTm1 (matched DP mechanism).}
    DP-vanilla and DP-aware are trained under matched DP-SGD settings ($C=1.0$, noise multiplier $\sigma = 0.1$, batch size $B=96$, and steps $T=20,000$).}
    \label{tab:main-results-cts}
\end{table*}

We repeat the main comparison on the ETTh1 and ETTm1 datasets. Tables~\ref{tab:main-results-bts} and~\ref{tab:main-results-cts} report representative metrics under matched DP mechanisms. Complete metric tables follow the same format as Table~\ref{tab:forecasting-full} and are omitted for brevity.

\subsection{Full Results on Effects of $\mathcal{B}_M(\cdot)$}\label{app:C3}

Here, we present additional results on the effect of the AdaLN modulation bounding operator $\mathcal{B}_M(\cdot)$ across tasks and noise levels. As shown in Tables \ref{tab:boundop-pp-forecast} and \ref{tab:boundop-pp-impute}, comparisons of bounding operators on PrivatePower for forecasting and interpolation/imputation tasks exhibit trends consistent with those in the main text: smooth operators ($\tanh$ and \texttt{soft\_clamp\_band}) achieve comparable performance and consistently outperform \texttt{hard\_clamp} and \texttt{clamp\_ste}.

\begin{table*}[t]
\centering
\resizebox{0.92\textwidth}{!}{
\begin{tabular}{lccc|ccc}
\toprule
& \multicolumn{3}{c}{$\sigma=0.05$} & \multicolumn{3}{c}{$\sigma=0.20$} \\
\cmidrule(lr){2-4}\cmidrule(lr){5-7}
$\mathcal{B}_M(\cdot)$
& point\_RMSE$\downarrow$ & dist\_JS$\downarrow$ & temp\_spec\_dist$\downarrow$
& point\_RMSE$\downarrow$ & dist\_JS$\downarrow$ & temp\_spec\_dist$\downarrow$ \\
\midrule
$\tanh$ (default)      & 0.423 & 0.636 & 8.558e-4 & 1.262 & 0.732 & 1.052e-2 \\
\texttt{soft\_clamp\_band} & 0.425 & 0.639 & 8.70e-4  & 1.270 & 0.737 & 1.070e-2 \\
\texttt{hard\_clamp}       & 0.452 & 0.662 & 9.45e-4  & 1.310 & 0.752 & 1.120e-2 \\
\texttt{clamp\_ste}        & 0.467 & 0.674 & 9.90e-4  & 1.330 & 0.760 & 1.150e-2 \\
\bottomrule
\end{tabular}}
\caption{\textbf{Bounding-operator comparison on PrivatePower for forecasting under matched DP-SGD.} Smooth operators ($\tanh$ and \texttt{soft\_clamp\_band}) perform similarly and outperform hard truncation variants (\texttt{hard\_clamp} and \texttt{clamp\_ste}).}
\label{tab:boundop-pp-forecast}
\end{table*}

\begin{table*}[t]
\centering
\resizebox{0.92\textwidth}{!}{
\begin{tabular}{lccc|ccc}
\toprule
& \multicolumn{3}{c}{$\sigma=0.05$} & \multicolumn{3}{c}{$\sigma=0.20$} \\
\cmidrule(lr){2-4}\cmidrule(lr){5-7}
$\mathcal{B}_M(\cdot)$
& point\_RMSE$\downarrow$ & point\_MAPE$\downarrow$ & point\_MAE$\downarrow$
& point\_RMSE$\downarrow$ & point\_MAPE$\downarrow$ & point\_MAE$\downarrow$ \\
\midrule
$\tanh$ (default)      & 2.019 & 4.97\% & 1.987 & 4.689 & 8.41\% & 3.442 \\
\texttt{soft\_clamp\_band} & 2.025 & 4.99\%  & 1.995 & 4.710 & 8.44\%  & 3.460 \\
\texttt{hard\_clamp}       & 2.120 & 5.22\%  & 2.070 & 4.860 & 8.70\%  & 3.560 \\
\texttt{clamp\_ste}        & 2.180 & 5.35\%  & 2.120 & 4.950 & 8.85\%  & 3.620 \\
\bottomrule
\end{tabular}}
\caption{\textbf{Bounding-operator comparison on PrivatePower for interpolation/imputation under matched DP-SGD.} Trends match forecasting: $\tanh$ and \texttt{soft\_clamp\_band} are consistently better than hard truncation variants.}
\label{tab:boundop-pp-impute}
\end{table*}

\section{Clipping Behavior under DP}
\label{app:clip-norm}

\begin{table*}[!ht]
    \centering
    \resizebox{0.68\textwidth}{!}{
    \begin{tabular}{lccccccc}
        \toprule
        Model & $\sigma$ &
        $p_{\mathrm{clip}} \downarrow$ &
        \makecell{mean($\eta$) $\uparrow$} &
        \makecell{p10($\eta$) $\uparrow$} &
        \makecell{p50($\eta$) $\uparrow$} &
        \makecell{p90($\eta$) $\uparrow$} &
        \makecell{p99($\eta$) $\uparrow$} \\
        \midrule
        DP-vanilla & 0.03 & 0.561   & 0.73   & 0.18   & 0.86   & 1.00   & 1.00   \\
        DP-aware   & 0.03 & 0.554   & \textbf{0.75}   & \textbf{0.22}   & \textbf{0.88}   & \textbf{1.00}   & \textbf{1.00}   \\
        \midrule
        DP-vanilla & 0.05 & 0.570   & 0.72   & 0.16   & 0.84   & 1.00   & 1.00   \\
        DP-aware   & 0.05 & 0.568   & \textbf{0.73}   & \textbf{0.18}   & \textbf{0.85}   & \textbf{1.00}   & \textbf{1.00}   \\
        \midrule
        DP-vanilla & 0.10 & 0.582   & 0.71   & 0.14   & 0.83   & 1.00   & 1.00   \\
        DP-aware   & 0.10 & 0.578   & \textbf{0.72}   & \textbf{0.16}   & \textbf{0.84}   & \textbf{1.00}   & \textbf{1.00}   \\
        \midrule
        DP-vanilla & 0.20 & 0.589   & 0.70   & 0.12   & 0.82   & 1.00   & 1.00   \\
        DP-aware   & 0.20 & 0.582   & \textbf{0.72}   & \textbf{0.15}   & \textbf{0.84}   & \textbf{1.00}   & \textbf{1.00}   \\
        \bottomrule
    \end{tabular}}
    \caption{\textbf{Clipping behaviour under DP-SGD on PrivatePower (matched DP mechanism).} We report the clipping activation rate $p_{\mathrm{clip}}$ and summary statistics of the clipping factor $\eta=\min(1,\frac{C}{\|g\|_2})$. Lower $p_{\mathrm{clip}}$ and larger typical $\eta$ indicate less clipping-induced attenuation.}
    \label{tab:clipping-rate}
\end{table*}

We report additional statistics characterizing the behavior of global per-sample clipping during DP-SGD training. These measurements complement the utility results by illustrating how clipping modifies the update signal. Given a clipping norm $C$, we define the clipping activation rate
\[
    p_{\mathrm{clip}} \coloneqq \Pr\!\big[\|g_i\|_2 > C\big],
\]
which we estimate from the per-sample gradients observed during training, as well as the clipping factor
\[
    \eta_i \coloneqq \min\!\left(1,\frac{C}{\|g_i\|_2}\right),
\]
whose distribution captures clipping severity, with smaller $\eta_i$ corresponding to stronger rescaling.

Table~\ref{tab:clipping-rate} reports $p_{\mathrm{clip}}$ and summary statistics of $\eta_i$ for DP-vanilla and DP-aware on PrivatePower across different noise multipliers $\sigma$. For each $\sigma$, we use the same DP mechanism across methods, so any differences reflect the impact of DP-aware bounds on gradient dynamics. Consistent with the main text, we observe comparable $p_{\mathrm{clip}}$ for the two approaches, while DP-aware yields slightly larger lower-quantile and median values of $\eta$ (i.e., milder rescaling), indicating reduced clipping severity.

\begin{figure*}[!ht]
    \centering
    \subfigure[\textbf{Clipping behavior at noise multiplier $\sigma=0.03$.}]{
        \includegraphics[width=0.485\textwidth]{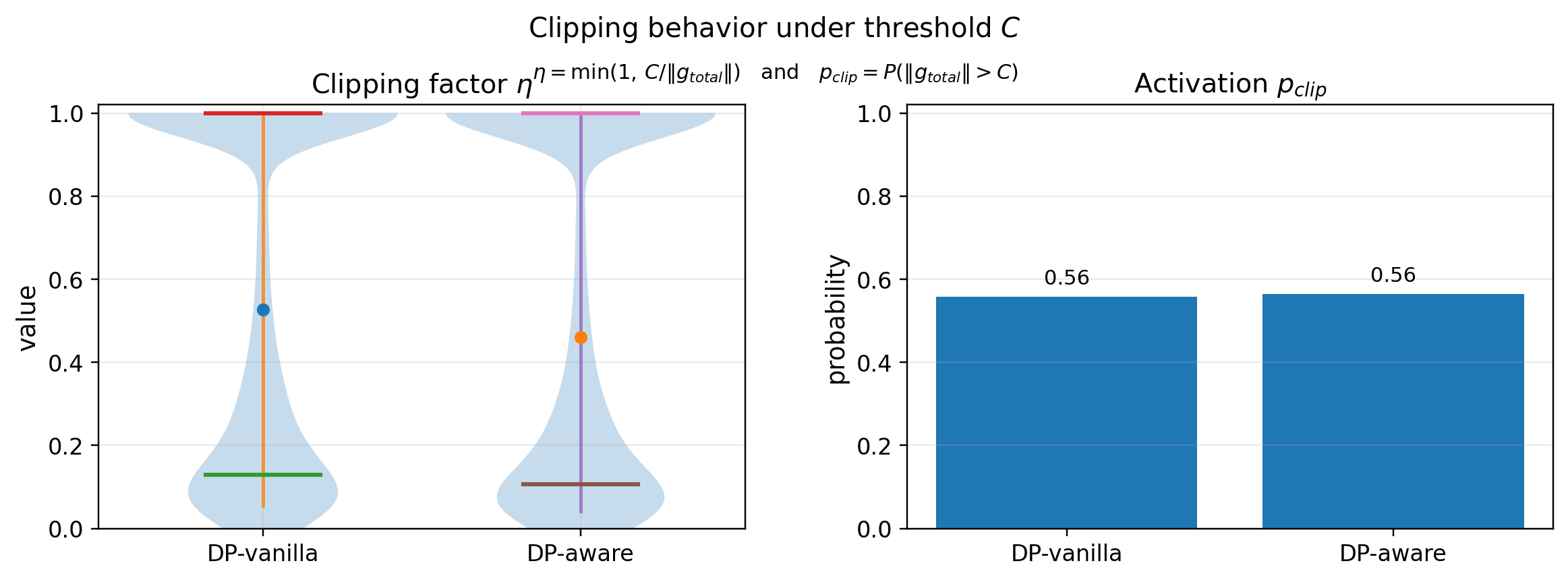}
    }
    \subfigure[\textbf{Clipping behavior at noise multiplier $\sigma=0.05$.}]{
        \includegraphics[width=0.485\textwidth]{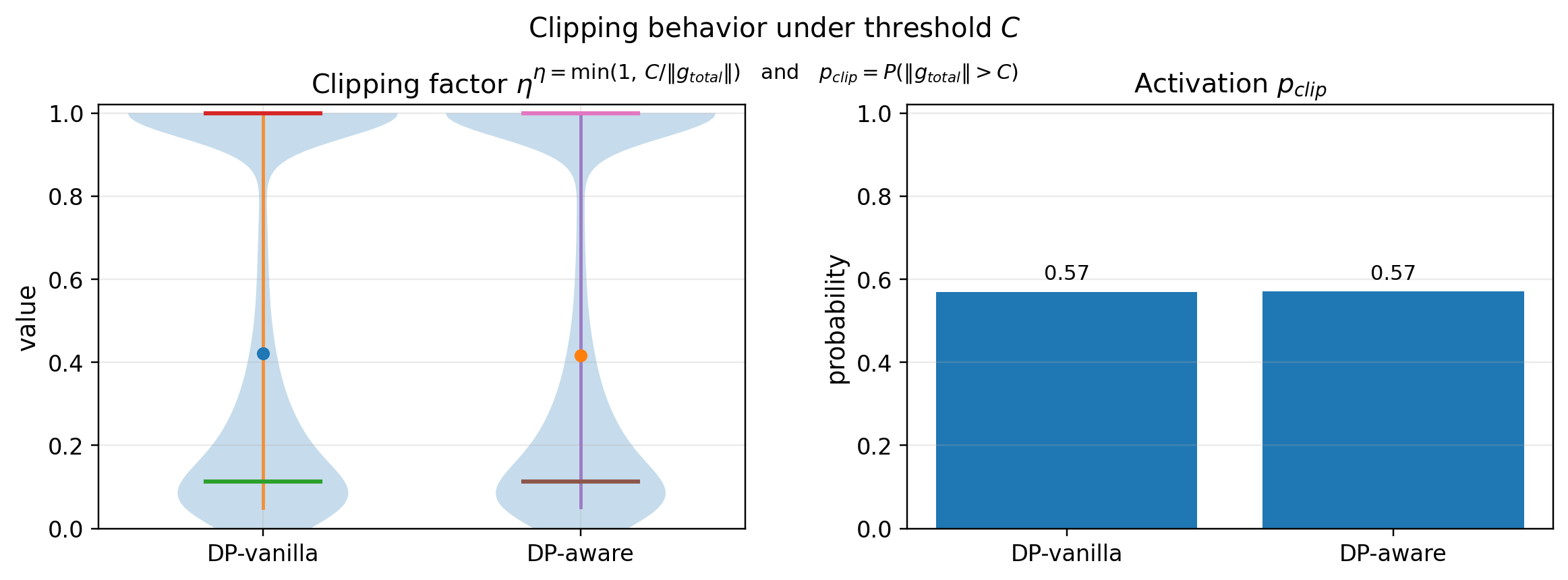}
    }
    \subfigure[\textbf{Clipping behavior at noise multiplier $\sigma=0.10$.}]{
        \includegraphics[width=0.485\textwidth]{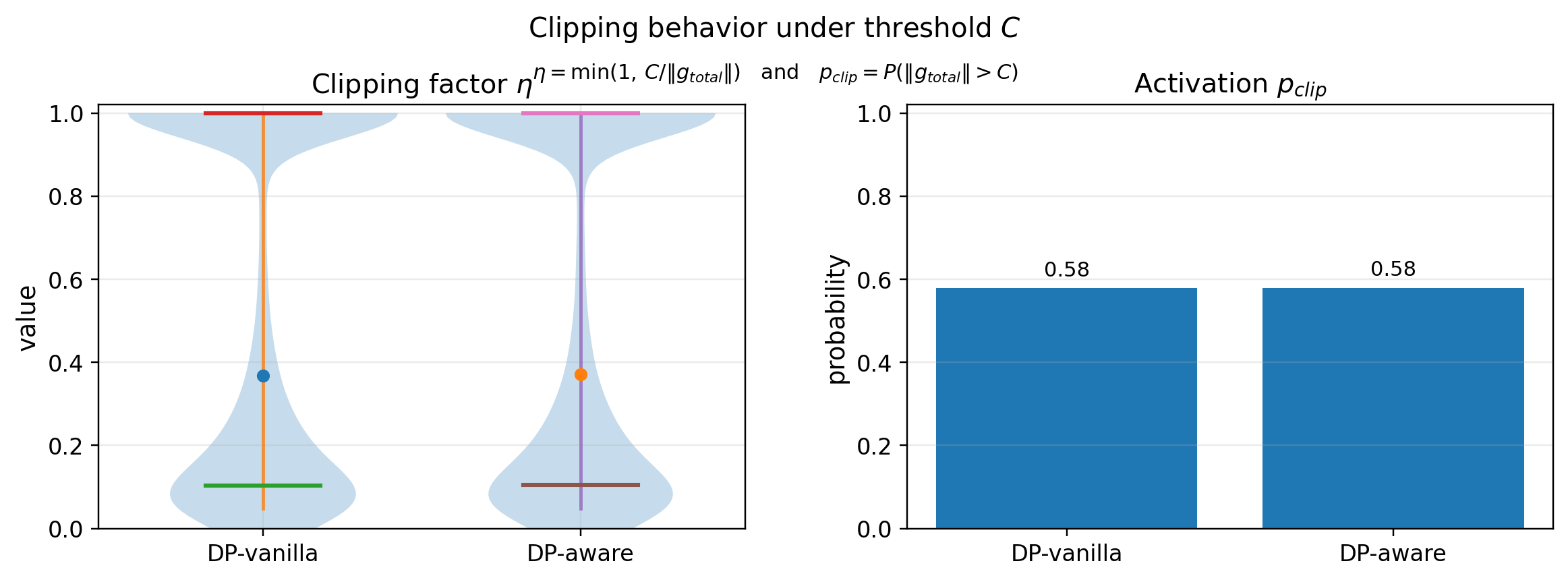}
    }
    \subfigure[\textbf{Clipping behavior at noise multiplier $\sigma=0.20$.}]{
        \includegraphics[width=0.485\textwidth]{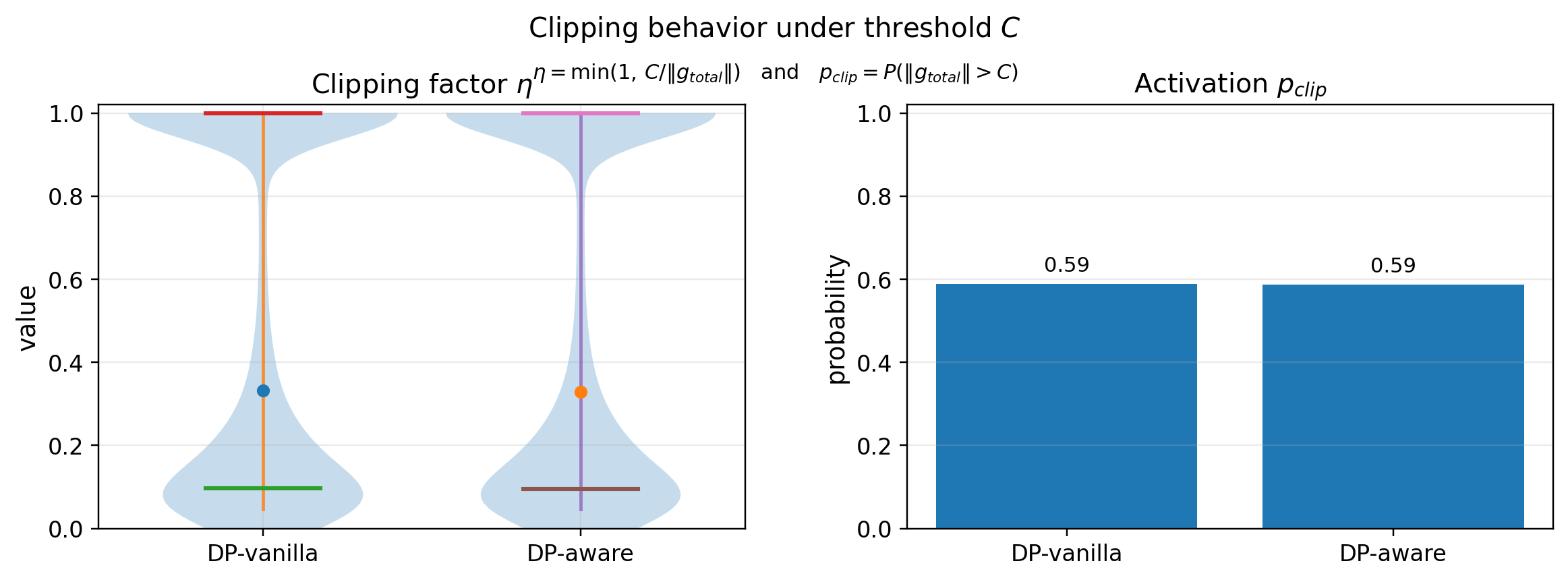}
    }

    \caption{\textbf{Clipping behavior under threshold $C$ (all $\sigma$ settings).} We compare DP-vanilla and DP-aware in terms of clipping factor $\eta=\min(1,\frac{C}{\|g_{\mathrm{total}}\|})$ and clipping activation rate $p_{\mathrm{clip}}=\mathbb{P}(\|g_{\mathrm{total}}\|>C)$.}
    \label{fig:clipping-behaviour-extended}
\end{figure*}

Furthermore, Figure~\ref{fig:clipping-behaviour-extended} visualizes the clipping behavior under the clipping threshold $C$ across all $\sigma$ settings. We compare DP-vanilla and DP-aware in terms of the clipping factor $\eta=\min(1,\, \frac{C}{\|g_{\mathrm{total}}\|})$ and the clipping activation rate $p_{\mathrm{clip}}=\mathbb{P}(\|g_{\mathrm{total}}\|>C)$. Across all noise levels, DP-aware exhibits a clipping frequency and a clipping-factor distribution comparable to DP-vanilla, consistent with the main-text conclusion that the improvements primarily arise from reshaping the gradient tail and reducing the update distortion, rather than from a lower clipping activation rate.

\begin{table*}[!ht]
    \centering
    \resizebox{0.72\textwidth}{!}{
        \begin{tabular}{lccccccc}
        \toprule
        Setting & $c_{\max}$ & $\gamma_{\max}$ & $\beta_{\max}$ & $\alpha_{\max}$ &
        $\rho_{\text{emp}}$ &
        point\_RMSE $\downarrow$ &
        dist\_JS $\downarrow$ \\
        \midrule
        \textit{Loose}     & $1.00\, C_0$  & $1.00\, \Gamma_0$  & $1.00\, B_0$  & $1.00\,A_0$  & 0.92   & 0.452   & 0.661   \\
        \textit{Medium}    & $0.90 \,C_0$  & $0.90\, \Gamma_0$  & $0.90\, B_0$  & $0.90\,A_0$  & 0.87   & \textbf{0.423}   & \textbf{0.636}   \\
        \textit{Tight}     & $0.75\, C_0$ & $0.75\, \Gamma_0$ & $0.75\, B_0$ & $0.75\,A_0$ & 0.80   & 0.441   & 0.651   \\
        \textit{Too tight} & $0.50\, C_0$  & $0.50\, \Gamma_0$  & $0.50\, B_0$  & $0.50\,A_0$  & 0.73   & 0.562   & 0.750   \\
        \bottomrule
        \end{tabular}
        }
    \caption{\textbf{Effect of DP-aware bound tightness on PrivatePower for forecasting.} DP-vanilla and all DP-aware variants use a matched DP-SGD mechanism (global clipping norm $C=1.0$, noise multiplier $\sigma=0.05$, batch size $B=96$, and training steps $T=20{,}000$). We scale a base reference setting $(C_0,\Gamma_0,B_0,A_0)$ for DP-aware bounds. We report $\rho_{\text{emp}}(p99)$ (average-case diagnostic; smaller indicates lighter tails/smaller typical gradients) and forecasting metrics (lower is better). The best performance is highlighted in bold.}
    \label{tab:tightness-forecast}
\end{table*}

\begin{table*}[!ht]
    \centering
    \resizebox{0.98\textwidth}{!}{
    \begin{tabular}{lccc|ccc}
        \toprule
        & \multicolumn{3}{c}{Interpolation/Imputation} &
          \multicolumn{3}{c}{Forecasting} \\
        \cmidrule(lr){2-4} \cmidrule(lr){5-7}
        Model &
        \makecell{point\_RMSE $\downarrow$} &
        \makecell{dist\_JS $\downarrow$} &
        \makecell{temp\_spectral\_dist $\downarrow$} &
        \makecell{point\_RMSE $\downarrow$} &
        \makecell{dist\_JS $\downarrow$} &
        \makecell{temp\_spectral\_dist $\downarrow$} \\
        \midrule
        Non-DP &
        0.584   & 0.66   & 1.48e-4   &
        0.208   & 0.77   & 1.48e-4   \\
        $\quad$+ DP-aware (\textit{Loose}) &
        0.590   & 0.67   & 1.52e-4   &
        0.212   & 0.62   & 1.55e-4   \\
        $\quad$+ DP-aware (\textit{Medium}) &
        \textbf{0.583}   & \textbf{0.66}   & \textbf{1.48e-4}   &
        \textbf{0.207}   & \textbf{0.77}   & \textbf{1.48e-4}   \\
        $\quad$+ DP-aware (\textit{Tight})&
        0.600   & 0.68   & 1.60e-4   &
        0.218   & 0.63   & 1.62e-4   \\
        $\quad$+ DP-aware (\textit{Too tight}) &
        0.720    & 0.78   & 2.30e-4   &
        0.290    & 0.75   & 2.40e-4   \\
        \bottomrule
    \end{tabular}}
    \caption{\textbf{Expressiveness check on PrivatePower without DP noise.} We compare a non-private baseline to non-private DP-aware variants under different tightness settings. Lower is better for all metrics shown. The best performance is highlighted in bold.}
    \label{tab:nonprivate-expressiveness}
\end{table*}


\section{Additional Ablations}
\label{app:ablations}

\subsection{Effect of Tightness Configurations.}
\label{app:tightness}
We sweep a small set of tightness configurations by scaling a base setting $(C_0,\Gamma_0,B_0,A_0)$: \textit{Loose}, \emph{Medium}, \emph{Tight}, and \emph{Too tight}. Here $(C_0,\Gamma_0,B_0,A_0)$ denote reference magnitudes for the conditioning pathway. We select these values via offline diagnostics to avoid trivial saturation while maintaining stable DP training. Table~\ref{tab:tightness-forecast} reports (i) the average-case diagnostic $\rho_{\text{emp}}$ and (ii) forecasting utility under matched DP hyperparameters. The \textit{Medium} setting typically performs best, providing sufficient tail control to stabilize DP training without overly restricting the conditioning pathway. As the bounds tighten, $\rho_{\text{emp}}$ decreases, reflecting stronger suppression of large gradients. Utility improves up to a point (\textit{Medium}) and then degrades when the bounds become overly restrictive (\textit{Too tight}). This behavior is consistent with the expected trade-off: suppress rare, conditioning-driven spikes without attenuating the conditioning signal.

\subsection{Effect of DP-aware Architectural Constraints}
\label{app:nonprivate_dpa}

This section isolates the effect of DP-aware architectural constraints from DP noise by evaluating DP-aware models trained without DP-SGD. We train (i) a non-private baseline and (ii) non-private DP-aware variants with DP-aware bounds enabled, but without per-sample gradient clipping and DP noise injection. We adopt the same tightness configurations as in Section~\ref{app:tightness} (\textit{Loose}, \textit{Medium}, \textit{Tight}, and \textit{Too tight}), and keep the architecture, optimizer, and training schedule fixed.

Table~\ref{tab:nonprivate-expressiveness} reports representative metrics on the PrivatePower dataset. Under the \textit{Medium} configuration, the non-private DP-aware variant matches the non-private baseline across metrics, suggesting that moderate bounds do not materially reduce model expressiveness. In contrast, the \textit{Too tight} configuration degrades performance, consistent with underfitting induced by overly restrictive bounds. Overall, these results support the interpretation that \textit{Medium} DP-aware bounds primarily suppress rare conditioning-driven outliers (which is beneficial under DP-SGD), without harming non-private performance.

\begin{table}[ht]
    \centering
    \resizebox{0.38\textwidth}{!}{
    \begin{tabular}{lccccc}
        \toprule
        $C$ &
        Model &
        $\rho_{\text{emp}}$ &
        \begin{tabular}[c]{@{}c@{}}
        RMSE $\downarrow$
        \end{tabular} &
        \begin{tabular}[c]{@{}c@{}}
        dist\_JS $\downarrow$
        \end{tabular} \\
        \midrule
        0.5 & DP-vanilla & 1.00 & 2.726   & 0.978   \\
        0.5 & DP-aware   & 0.87 & \textbf{1.145}   & \textbf{0.826}   \\
        \midrule
        1.0 & DP-vanilla & 1.00 & 1.673   & 0.833   \\
        1.0 & DP-aware   & 0.89 & \textbf{0.671}   & \textbf{0.757}   \\
        \midrule
        2.0 & DP-vanilla & 1.00 & 1.344   & 0.787   \\
        2.0 & DP-aware   & 0.89 & \textbf{0.446}   & \textbf{0.664}   \\
        \bottomrule
    \end{tabular}}
    \caption{\textbf{Effect of clipping threshold $C$ on PrivatePower for forecasting at $\sigma = 0.1$.} All rows use a matched DP mechanism for each $C$ (same $\sigma$, $B$, and $T$); only $C$ changes. DP-aware consistently outperforms DP-vanilla across a range of $C$ values.}
    \label{tab:clip-norm}
\end{table}

\subsection{Effect of Clipping Threshold}

We also study the effect of the clipping threshold \(C\) in DP-SGD. Table~\ref{tab:clip-norm} reports representative forecasting metrics and the empirical diagnostic \(\rho_{\text{emp}}\) for several values of \(C\) under a matched DP mechanism. We find that increasing $C$ consistently improves utility: as $C$ grows, both RMSE and dist\_JS decrease for DP-vanilla and DP-aware, indicating better forecasting utility and a closer predictive distribution to the reference. Meanwhile, $\rho_{\text{emp}}$ remains at $1.00$ for DP-vanilla and stays stable around $0.87-0.89$ for DP-aware. Across a wide range of $C$, DP-aware consistently outperforms DP-vanilla.

\subsection{Effects of Interpolation/Imputation Metrics}
\label{app:ablations-interp}

Table~\ref{tab:tightness-interp} mirrors the forecasting tightness ablation in the main text, but reports interpolation/imputation metrics. All rows use a matched DP mechanism; we vary only the DP-aware bound tightness. The same ``sweet spot'' persists for interpolation: moderately tight bounds improve utility, whereas overly restrictive bounds lead to underfitting. Table~\ref{tab:component-ablation-interp} extends the component-wise ablation to interpolation/imputation. Each row uses the same DP mechanism; only the inclusion of the two DP-aware components differs.

\begin{table}[!ht]
    \centering
    \resizebox{0.32\textwidth}{!}{
    \begin{tabular}{lccc}
        \toprule
        Setting &
        $\rho_{\text{emp}}$ &
        \begin{tabular}[c]{@{}c@{}}
        RMSE $\downarrow$
        \end{tabular} &
        \begin{tabular}[c]{@{}c@{}}
        MAE $\downarrow$
        \end{tabular} \\
        \midrule
        \textit{Loose}     & 0.92 & 2.231   & 2.170   \\
        \textit{Medium}    & 0.87 & \textbf{2.019}   & \textbf{1.987}   \\
        \textit{Tight}     & 0.80 & 2.672   & 2.246   \\
        \textit{Too tight} & 0.73 & 3.626   & 3.179   \\
        \bottomrule
    \end{tabular}}
    \caption{\textbf{Effect of DP-aware bound tightness on PrivatePower for interpolation/imputation at $\sigma = 0.05$ under a matched DP mechanism.} We report $\rho_{\text{emp}}$ and representative interpolation metrics (lower is better); boldface denotes the best performance.}
    \label{tab:tightness-interp}
\end{table}

\begin{table}[!ht]
    \centering
    \resizebox{0.52\textwidth}{!}{
    \begin{tabular}{lccc}
        \toprule
        Model variant &
        $\rho_{\text{emp}}$ &
        \begin{tabular}[c]{@{}c@{}}
        RMSE $\downarrow$
        \end{tabular} &
        \begin{tabular}[c]{@{}c@{}}
        MAE $\downarrow$
        \end{tabular} \\
        \midrule
        DP-vanilla                                 & 1.00 & 3.498   & 3.110   \\
        DP-aware (only $\mathbf{c}$ bounded)       & 0.95 & 3.998   & 3.070   \\
        DP-aware (only AdaLN bounded)              & 0.93 & 2.337   & 2.112   \\
        DP-aware (full, ours)                      & 0.89 & \textbf{2.019}   & \textbf{1.987}   \\
        \bottomrule
    \end{tabular}}
    \caption{\textbf{Effects of forward-pass control components on PrivatePower for interpolation/imputation at $\sigma=0.05$.} All rows use a matched DP mechanism. We report the empirical diagnostic $\rho_{\text{emp}}$ and representative interpolation metrics (lower is better). Boldface indicates the best performance.}
    \label{tab:component-ablation-interp}
\end{table}

\section{Additional Training-Time Signals}
\label{app:training-dynamics}

This section reports additional training-time signals under DP-SGD, including: (i) training-loss trajectories across noise multipliers and (ii) representative wall-clock step time. All results are evaluated on PrivatePower using the same optimizer and training schedule as in the main experiments.

\subsection{Training Loss across Noise Multipliers}
\label{app:training-loss}

\begin{figure*}[ht]
    \centering
    \subfigure[Training MSE loss ($\sigma=0.03$).]{
        \includegraphics[width=0.485\textwidth]{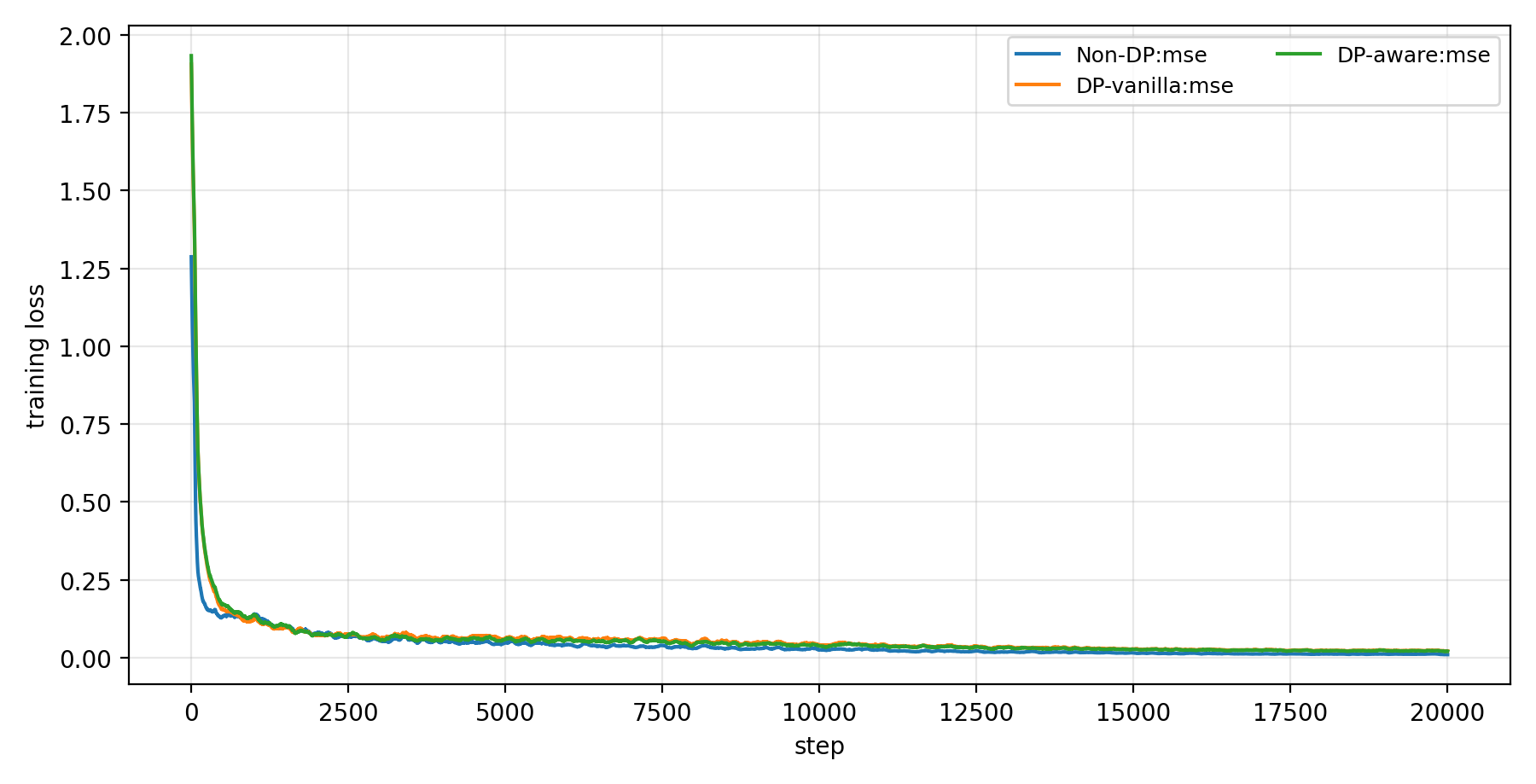}
        \label{fig:mseloss:0.001}
    }
    \subfigure[Training MSE loss ($\sigma=0.05$).]{
        \includegraphics[width=0.485\textwidth]{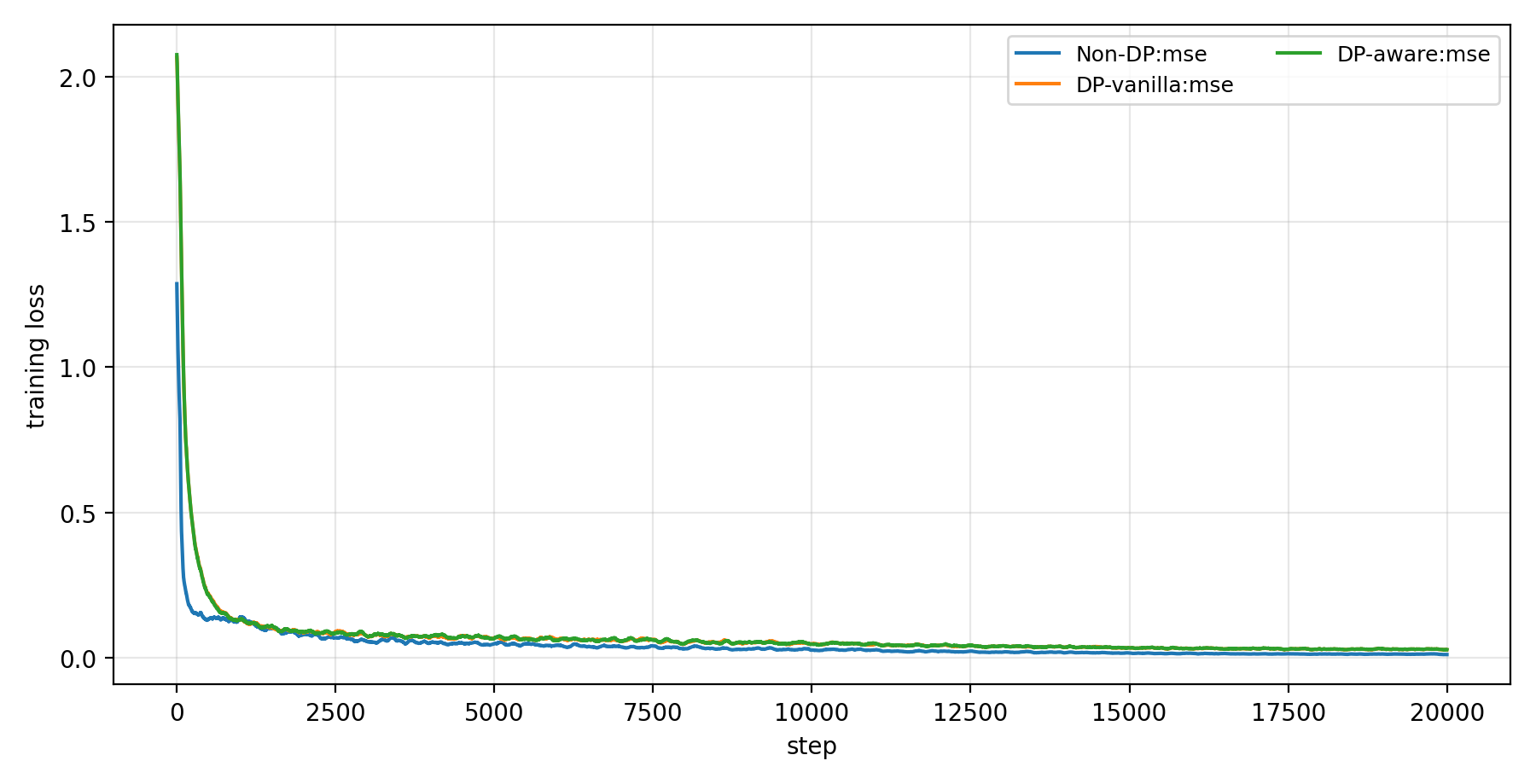}
        \label{fig:mseloss:0.003}
    }
    \subfigure[Training MSE loss ($\sigma=0.1$).]{
        \includegraphics[width=0.485\textwidth]{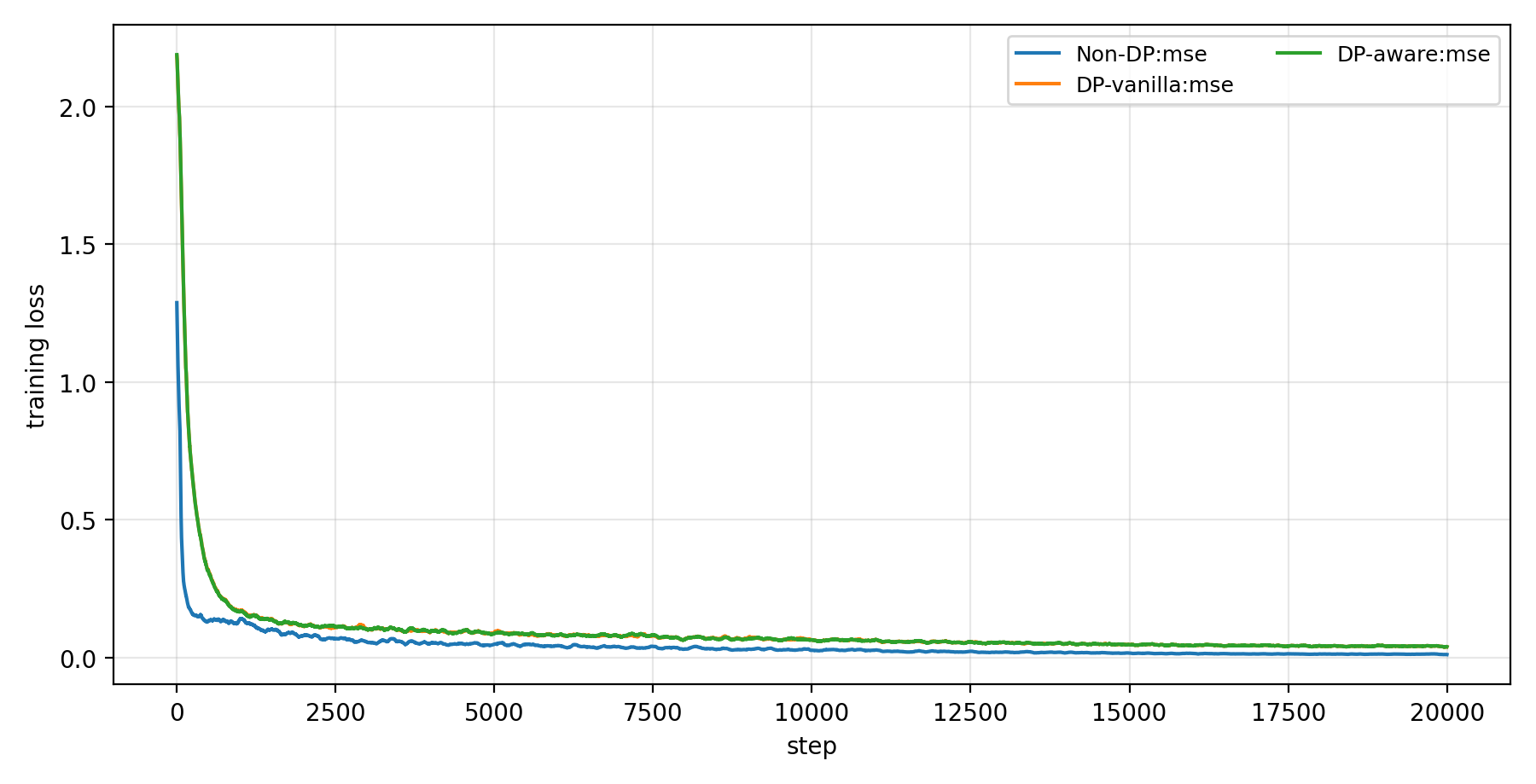}
        \label{fig:mseloss:0.007}
    }
    \subfigure[Training MSE loss ($\sigma=0.2$).]{
        \includegraphics[width=0.485\textwidth]{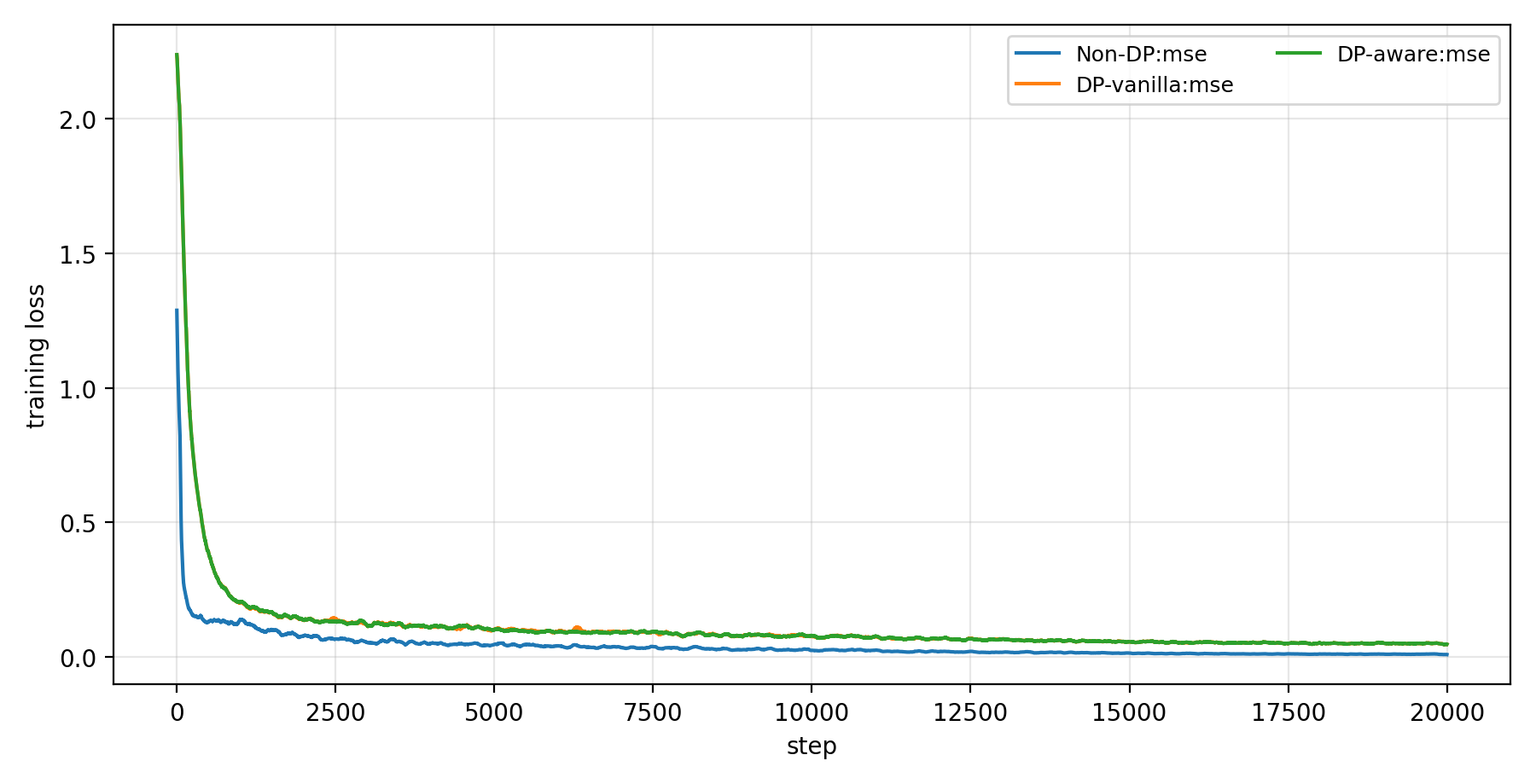}
        \label{fig:mseloss:0.01}
    }
    \caption{\textbf{Training loss dynamics on PrivatePower.} Training MSE loss curves for Non-DP, DP-vanilla, and DP-aware under several DP noise multipliers. Larger $\sigma$ yields a higher loss floor, while DP-aware closely follows DP-vanilla across noise levels.}
    \label{fig:training-loss-curves}
\end{figure*}

Figure~\ref{fig:training-loss-curves} plots the evolution of the training MSE loss over optimization steps for Non-DP, DP-vanilla, and DP-aware runs under several noise multipliers $\sigma \in \{0.03, 0.05, 0.1, 0.2\}$. As expected, increasing the DP noise scale raises the asymptotic loss, with larger $\sigma$ leading to a higher steady-state loss floor. Across all noise settings, DP-aware tracks DP-vanilla closely, suggesting that our proposed architectural constraints do not adversely affect optimization or training stability. In contrast, the Non-DP baseline attains the lowest final loss, consistent with the absence of gradient clipping effects and DP noise.

\subsection{Runtime: Step Time under DP Training}
\label{app:runtime}

We additionally report an average per-step wall-clock time comparison over 5 runs. Table~\ref{tab:runtime-step-time} summarizes robust statistics of the step time.

\begin{table}[!ht]
    \centering
    \resizebox{0.44\textwidth}{!}{
    \begin{tabular}{lcccc}
        \toprule
        Method & mean $\downarrow$ & p50 $\downarrow$ & p90 $\downarrow$ & rel.\ to Non-DP \\
        \midrule
        Non-DP     & 0.100   & 0.047   & 0.201   & 1.00$\times$ \\
        DP-vanilla & 0.168   & 0.105   & 0.350   & 1.68$\times$ \\
        DP-aware   & 0.169   & 0.108   & 0.351   & 1.69$\times$ \\
        \bottomrule
    \end{tabular}}
    \caption{\textbf{Runtime summary.} Robust statistics of per-step wall-clock time for Non-DP, DP-vanilla, and DP-aware training.}
    \label{tab:runtime-step-time}
\end{table}


\section{Details for Datasets and Mask-Based Tasks}
\label{app:exp-details}

Table~\ref{tab:datasets} provides the configurations of the datasets and their corresponding mask-based tasks.

\begin{table*}[t]
    \centering
    \resizebox{0.99\textwidth}{!}{
    \begin{tabular}{lcccccccc}
        \toprule
        Dataset & Records & Channels & Seq Len & Train/Val/Test & Normalization &
         \texttt{ratio\_range} &
         \texttt{pred\_len\_range} &
         \texttt{num\_blocks\_range} \\
        \midrule
        PrivatePower &
        $\approx$23.1k &
        7 &
        168 &
        70\%/15\%/15\% &
        per-channel standardization &
        [0.1, 0.5] &
        [24, 96] &
        [4, 8] \\
        ETTh1 &
        $\approx$17.5k &
        7 &
        96 &
        12m/4m/4m &
        per-feature z-score   &
        [0.1, 0.5] &
        [24, 96] &
        [4, 8] \\
        ETTm1 &
        $\approx$70k &
        7 &
        96 &
        12m/4m/4m  &
        per-feature z-score &
        [0.1, 0.5] &
        [24, 96] &
        [4, 8] \\
        \bottomrule
    \end{tabular}}
    \caption{\textbf{Configurations of the datasets and their corresponding mask-based tasks.} Each dataset trains a single conditional diffusion model that supports interpolation/imputation (random \& stride masks), forecasting (block masks), and full-sequence reconstruction. For PrivatePower, we follow the private-data pipeline in our codebase; for ETTh1 and ETTm1, we follow the common ETT setting.}
    \label{tab:datasets}
\end{table*}

\paragraph{PrivatePower.}
PrivatePower contains $23{,}136$ records with $7$ channels. We use $(K_{\max},L_{\max})=(7,168)$, split the data chronologically into train/validation/test ($70/15/15$), and apply per-channel standardization using training-split statistics (mean and standard deviation). The raw timestamp-indexed hourly \texttt{power\_usage} series (single meter) is preprocessed via an internal pipeline: (i) replace negative values by interpolation using a $7$-day moving-average baseline, and (ii) add calendar and temporal covariates derived from timestamps, followed by (iii) the above normalization. Training windows are constructed using a sliding window of length $168$ with stride $24$. For mask-based tasks, we use the ranges in Table~\ref{tab:datasets}: \texttt{ratio\_range} $[0.1,0.5]$, \texttt{pred\_len\_range} $[24,96]$, and \texttt{num\_blocks\_range} $[4,8]$. We train a conditional diffusion model with $1000$ diffusion steps (linear $\beta$ schedule) and a Transformer backbone (depth $8$, hidden size $256$, $8$ attention heads), using AdamW with learning rate $7\times10^{-4}$, weight decay $2\times10^{-5}$, batch size $96$, EMA decay $0.999$, for $20$k optimization steps.

\paragraph{ETTh and ETTm1.}
ETTh1 (hourly; $\approx$17.5k records) and ETTm1 ($15$-minute; $\approx$70k records) are public ETT benchmarks~\cite{zhou2021informer}. Both datasets have $7$ channels, and we use sequence length $L_{\max}=96$ with $K_{\max}=7$ to match the standard ETT window. We follow the standard chronological split ($12/4/4$ months for train/validation/test) and apply per-feature $z$-score normalization using training statistics. For mask-based tasks, we use the same ranges as in Table~\ref{tab:datasets}: \texttt{ratio\_range} $[0.1,0.5]$, \texttt{pred\_len\_range} $[24,96]$, and \texttt{num\_blocks\_range} $[4,8]$. For imputation-oriented masks, we use length-$96$ windows to align with common ETT evaluation practice~\cite{cao2024timedit}. We train one model per dataset from scratch, using the same model family and optimizer settings as for PrivatePower for a controlled comparison. To account for the larger number of available windows in ETTm1, we fix the total number of optimization steps and use random window sampling to provide training diversity.



\end{document}